\documentclass{article}

\usepackage[final]{neurips_2019}

\usepackage[utf8]{inputenc} 
\usepackage[T1]{fontenc}    

\usepackage{url}            
\usepackage{booktabs}       
\usepackage{amsfonts}       
\usepackage{nicefrac}       
\usepackage{microtype}      
\usepackage{subfig}
\usepackage{graphicx}
\usepackage{amsmath}
\usepackage{algorithm}
\usepackage[noend]{algpseudocode}

\newcommand{\mc}[1]{\mathcal{#1}}
\newcommand{\mb}[1]{\mathbb{#1}}

\usepackage{xcolor}
\usepackage{tikz}
\usepackage{enumerate}

\usepackage[shortlabels]{enumitem}

\usepackage{bbold}
\usepackage{xcolor,colortbl}
\usepackage{adjustbox}
\usepackage{color}
\usepackage{stackengine}
\definecolor{Gray}{gray}{0.85}
\newtheorem{definition}{Definition}

\title{Design, Benchmarking and Explainability Analysis\\of a Game-Theoretic Framework towards Energy Efficiency in Smart Infrastructure}

\author{%
Ioannis C. Konstantakopoulos\textsuperscript{\textdagger}\thanks{Corresponding Author. Email: \tt ioanniskon@berkeley.edu}, Hari Prasanna Das\textsuperscript{\textdagger}, Andrew R. Barkan\textsuperscript{\ddag}, Shiying He\textsuperscript{\textdagger},\\
\textbf{Tanya Veeravalli\textsuperscript{\textdagger}, Huihan Liu\textsuperscript{\textdagger}, Aummul Baneen Manasawala\textsuperscript{\textsection}, Yu-Wen Lin\textsuperscript{\textdagger}, Costas J. Spanos\textsuperscript{\textdagger}}\\
\textsuperscript{\textdagger}Department of Electrical Engineering and Computer Sciences, University of California, Berkeley\\
\textsuperscript{\ddag}Department of Mechanical Engineering, University of California, Berkeley\\
\textsuperscript{\textsection}Department of Industrial Engineering and Operations Research, University of California, Berkeley\\
}

\begin{document}

\maketitle
\begin{abstract}
In this work, we propose a gamification approach as a novel framework for smart building infrastructure with the goal of motivating human occupants to reconsider personal energy usage and to have positive effects on their environment. Human interaction in the context of cyber-physical systems is a core component and consideration in the implementation of any smart building technology. Research has shown that the adoption of human-centric building services and amenities leads to improvements in the operational efficiency of these cyber-physical systems directed towards controlling building energy usage. We introduce a strategy in form of a game-theoretic framework that incorporates humans-in-the-loop modeling by creating an interface to allow building managers to interact with occupants and potentially incentivize energy efficient behavior. Prior works on game theoretic analysis typically rely on the assumption that the utility function of each individual agent is known a priori. Instead, we propose novel utility learning framework for benchmarking that employs robust estimations of occupant actions towards energy efficiency. To improve forecasting performance, we extend the utility learning scheme by leveraging deep bi-directional recurrent neural networks. Using the proposed methods on data gathered from occupant actions for resources such as room lighting, we forecast patterns of energy resource usage to demonstrate the prediction performance of the methods. The results of our study show that we can achieve a highly accurate representation of the ground truth for occupant energy resource usage. We also demonstrate the explainable nature on human decision making towards energy usage inherent in the dataset using graphical lasso and granger causality algorithms. For demonstrations of our infrastructure and for downloading de-identified, high-dimensional data sets, please visit our website \footnote{Open sourced high-dimensional data and demonstrations: \url{https://smartntu.eecs.berkeley.edu}}. 
\end{abstract}
\section{Introduction}

\begin{figure}[t]
    \subfloat[Gamification abstraction for \textit{Human-Centric Cyber-Physical Systems}]{%
    \includegraphics[width=0.45\textwidth]{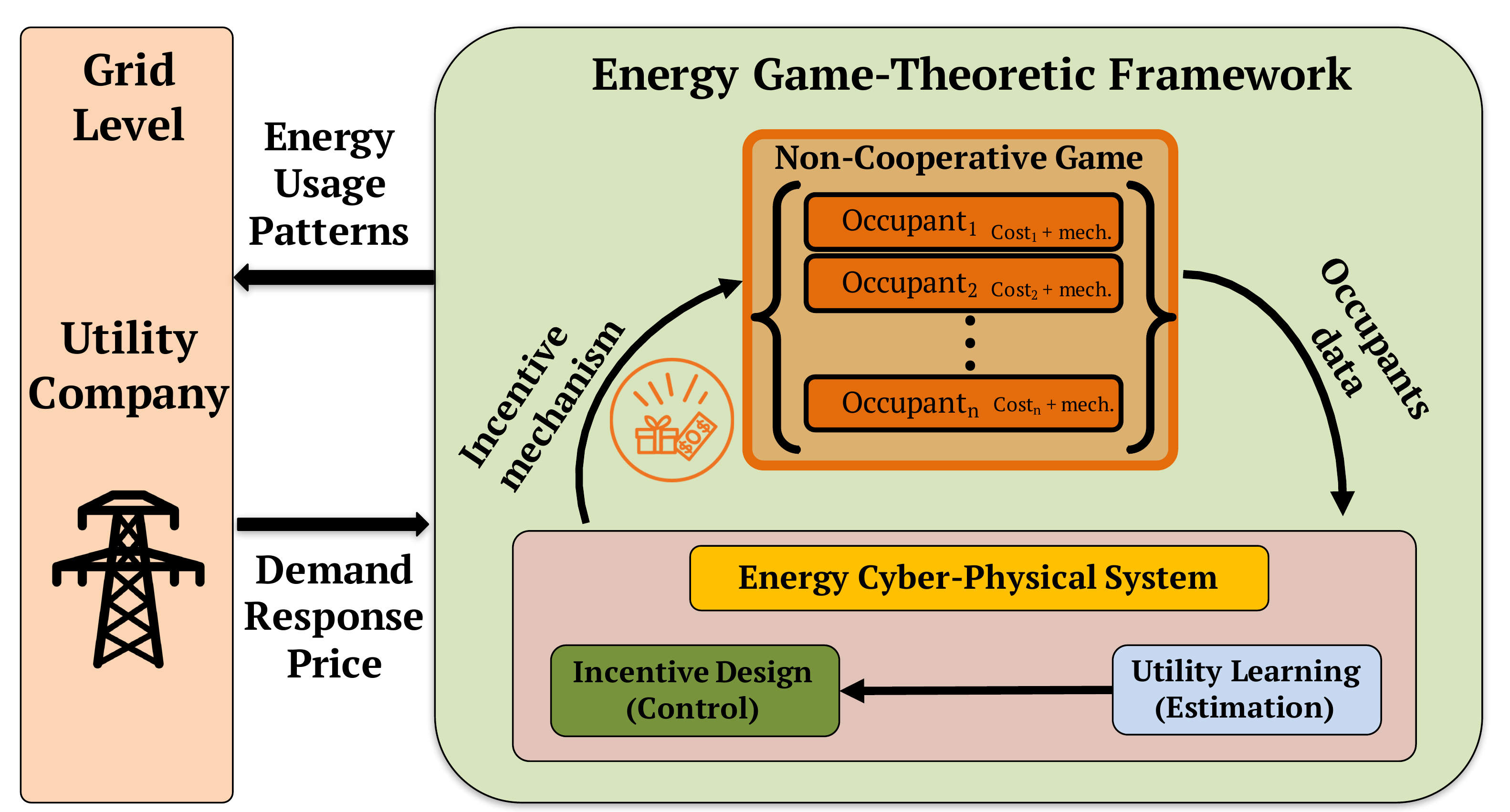}\label{fig:gamification_abstraction_A}
    }
    \hfill
    \subfloat[High level view of the proposed design framework]{%
    \includegraphics[width=0.48\textwidth]{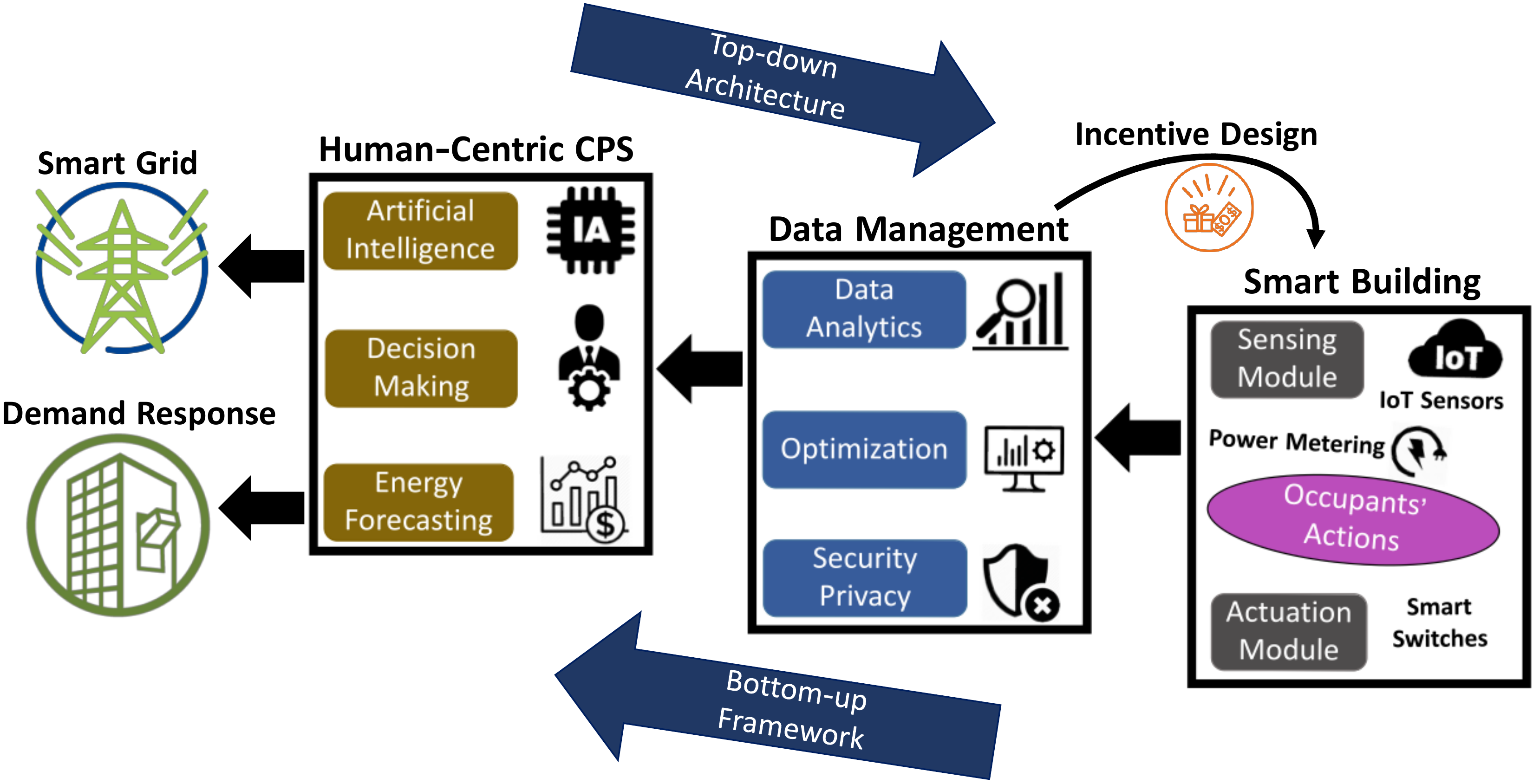}\label{fig:gamification_abstraction_B}
    }
\caption{Gamification abstraction and high level view of proposed framework}
\label{fig:gamification_abstraction_and_block_diagram}
\end{figure}

Nearly half of all energy consumed in the U.S. can be attributed to usage by residential and commercial buildings~\citep{mcquade2009}. In efforts to improve energy efficiency in buildings, researchers and industry leaders have attempted to implement novel control and automation approaches alongside techniques like incentive design and adaptive price adjustment to more effectively regulate energy usage. All the solutions pertaining to making a building occupant friendly and energy efficient fall under the umbrella of smart building techniques. To summarize the space, an ideal smart building infrastructure accommodates a variety of occupant preferences including thermal comfort~\citep{karmann2018percentage,Liu_thermal_comfort,liu2019personal}, satisfaction/well-being~\citep{frontczak2012quantitative}, lighting comfort~\citep{zhu2017illuminance}, acoustical quality~\citep{yang2017perceptions}, occupancy based control~\citep{zou2019machine}, indoor air quality~\citep{sundell2011ventilation}, indoor environmental monitoring~\citep{jin2018automated}, activity/gesture based control~\citep{zou2019wifi,zou2019consensus}, privacy~\citep{jia2017privacy} and productivity~\citep{horr2016environment}, while simultaneously optimizing for energy efficiency and agile connectivity to the grid. A common avenue for the regulation of energy usage in buildings is through their on-site building and facility managers. Building managers are obligated to maintain an energy efficient schedule according to a standard operating procedure. Even with these considerations, there is still a tremendous need for scalable and robust frameworks that can efficiently coordinate and control building resource usage in the presence of confounding dynamics such as human behavior. 


Recently, utility companies have invested in demand response programs that can address improper load forecasting while also helping building managers encourage energy efficiency among building occupants~\citep{shariatzadeh2015demand,bianchini2016demand}. Commonly, the implementation of these programs is enacted on a contract basis between utility providers and the consumers under arranged conditions of demand/usage. The building managers will then be bound by contract to operate according to the agreed-upon schedule. However, the conditions of these contracts are static and do not consider dynamic changes in occupant behavior or preferences, which can result in discrepancies in demand/usage expectations. To facilitate the adoption of more dynamic protocols for demand response, our setup features a gamification interface (seen in the building level in Figure~\ref{fig:gamification_abstraction_A}) that allows building managers to interact with a building's occupants. By leveraging our gamification interface, retailers and utility companies at the provider level can utilize a wealth of dynamic and temporal data on building energy usage, extending even to occupant usage predictions, in order to customize demand response program approaches to observed or predicted conditions~\cite{jin2017mod,jin2018microgrid}. Above all, our gamification interface is designed to support engagement and integration on multiple levels in a \textit{human-centric cyber-physical system}, which is formally defined as:


\begin{center}
\textit{Systems or mechanisms that combine computer-based technologies with physical processes to integrate direct human coordination through various interaction modalities.}
\end{center}

The availability of a game-theoretic framework will allow researchers to demonstrate gaming data in a discrete choice setting, run simulations that include data on dynamic occupant preferences, test correlations between actions and external parameters (e.g. provided weather metrics), and leverage temporal data in demand response program scenarios.

The cooperation of human agents with building automation in smart infrastructure setting helps in improving the system robustness and sustainability by taking advantage of both the control potential of computational methods and the real life information provided by humans-in-the-loop elements. The inherent adaptability and simultaneous automation in such a system makes it possible to accommodate a wide variety of dynamic situations that might arise in the maintenance of building infrastructure, like the automatic shifting of demand during peak energy/usage hours. Put into more broad terms, the goal of many building infrastructure systems is to enact system-level efficiency improvements by using a high-level \emph{planner} (e.g. facility manager) to coordinate autonomously acting agents in the system (e.g. selfish human decision-makers). It is this type of functionality that makes smart building technology so essential to the development of an ideal \emph{smart city}.


Our approach to efficient building energy management focuses on improving energy usage behavior among occupants in residential dorm buildings, by utilizing cutting-edge Internet of Things (IoT) sensors and cyber-physical system sensing/actuation platform integrated with the aforementioned gamification interface (Figure~\ref{fig:gamification_abstraction_A}). The interface is designed to support occupant engagement and integration while learning occupant preferences over shared resources. It also provides an window to understand how human preferences change as a function of critical factors such as manual control of devices, time of day, and provided incentives. Our gamification framework can be used in the design of incentive mechanisms in the form of fair compensation that help to realign agent preferences with those of the planner, which are often representative of system-level performance criteria.


We present a social game aimed at incentivizing occupants to modify their behavior so that the overall energy consumption in their room is reduced. In certain living situations, occupants in residential buildings are not responsible for paying for the energy resources they consume. For this reason, there is often an imbalance between the motivations of the facility manager and those of the occupants. The competitive aspects of the social game framework motivate occupants to address their inefficiencies while encouraging responsible energy usage on an individual basis. At the core of our approach is a decision model, which treats building occupants as non-cooperative agents who play according to a sequential discrete choice game. Discrete choice models have been used extensively to investigate representations for scenarios like variation in modes of transportation~\citep{qin2017heterogeneity}, demand for organic foods~\citep{hillier2015discrete}, and even school social interactions~\citep{karmargianni2014social}. Our framework is centered around learning agent preferences over room resources, such as lighting, as well as external parameters like weather conditions, high-level grid control, and provided incentives. Specifically, agents are strategic entities that make decisions based on their own preferences without consideration of other decision makers. The game-theoretic framework allows for qualitative insights to be made about the outcome of this selfish behavior (better than a simple predictive model) and, more importantly, can be leveraged in designing effective incentive mechanisms to motivate agents.

The broader motivation of this paper is to introduce a general framework that utilizes game theoretic concepts to learn models of players decision-making in residential buildings, which is made possible by the implementation of our energy game-theoretic framework. We present a variety of benchmark utility learning models and a novel pipeline for the efficient training of these models. In order to boost predictive power, we propose end-to-end Deep Learning models focused on the utilization of deep recurrent neural networks for analyzing gaming data. To handle sequential information dependencies in our data, we implement a deep learning based architecture using Long Short Term Memory cells (LSTM). With the advent of explainable Artificial Intelligence (AI), there has been a massive move towards making statistical models explainable. So, we perform feature correlation study using graphical lasso algorithm and causality study using grangers causality to confirm that the data obtained from the social game holds accurate information on human decision making towards energy usage in competetive settings. Finally, we open source our data and provide a web portal for demonstrating our infrastructure and for downloading de-identified, high-dimensional data\footnote{Demo web portal and open sourced data: \url{https://smartntu.eecs.berkeley.edu}}. High-dimensional data can serve either as a benchmark for alternative discrete choice model learning schemes or as a helpful source for analyzing occupant energy usage in residential buildings.



The body of the paper is organized as follows. Previous works are surveyed in Section~\ref{related_work} with an emphasis on human-centric models with integration in smart grid infrastructures. Section~\ref{social_game} describes the social game experiment on the Nanyang Technological University campus and the human decision-making model. Section~\ref{machine_learning} introduces utility estimation along with several proposed machine learning \& deep learning algorithms for sequential decision games. It also describes the methodology for explainability analysis. Results are presented in Section~\ref{results}. We make concluding remarks in Section~\ref{conclusion}.

\section{Related Work}
\label{related_work}



Smart grid technology focuses on enabling efficient grid integration and comprehensive analysis to support advances in renewable energy sources, power systems management, minimization of inefficiencies in usage, and maximization of user savings. However, challenges in power grid applications, such as the lack of predictability, and the stochastic environment in which the power grid operates complicate the synthesis of an all-encompassing solution. To address these problems, industry leaders and researchers in the fields of power grid design and control have put forth considerable research and development efforts in smart grid design, demand-side management, and power system reliability. 


With the help of digital hardware and information technology, smart grid design relies more and more on the development of decision-capable intelligence in the context of grid automation. Novel methods~\citep{salehi2012laboratory,das2016development} for smart grid design incorporate real-time analysis and stochastic optimization to provide power grid observability, controllability, security, and an overall reduction of operational costs. Specifically, the integration of data analytics and innovative software platforms have led to effective trends in demand-side management \citep{ramchurn2011agent} and demand response \citep{maharjan2013dependable}. These studies explored and drew upon methods from behavioral decision research to analyze the decision patterns of users and consumers. The simulations and empirical results from these studies reinforce the significance of forecasting energy demands and the potential advantages of managing these demands by leveraging models of intelligent decision-makers. In this context, we can see that the process of modeling and predicting the actions of decision-makers in the control of large networks is a significant development towards improving the operational efficiency of smart grids.




Game theory can serve as an extremely useful tool for the real-time forecasting of decision-makers in an interactive setting. Classical models from game theory allow for qualitative insights about the outcome of scenarios involving the selfish behavior of competitive agents and can be leveraged in the design of incentives for influencing the goals of these agents. Contemporary research in the energy and power systems domain leverage game theoretic models in a multitude of applications. As previously mentioned, these types of approaches have been implemented in the modeling of various aspects of smart grid control. Specifically, we can observe instances of game theory applications in the context of smart grid demand response programs using methods such as supply-balancing~\citep{yu2016supply}, hierarchical Stackelberg game settings~\citep{yu2017incentive}, and Vickrey-Clarke-Groves (VCG) auction mechanisms~\citep{samadi2012advanced}. The use of game theoretic models creates new avenues for modeling dynamic economic interactions between utility providers and consumers inside a distributed electricity market~\citep{zhang2015game}. Another example study is the investigation of crowdfunding as an incentive design methodology for the construction of electric vehicle charging piles~\citep{zhu2017study}. Game theory has also been directed towards the optimal design of curtailment schemes that control the fair allocation of curtailment among distributed generators~\citep{andoni2017game}. Expanding on previous work, researchers have studied game theory applications in the context of incentive-based demand response programs for customer energy reduction as well~\citep{vuelvas2018limiting}. In these types of applications, customer interaction with the incentive provider is modeled using game theory while their engagement is represented probabilistically.



In majority of the previously discussed game theoretic applications, results are generated purely by simulation without the use of real data. Furthermore, previous applications fail to propose any novel techniques for learning the underlying utility functions that dynamically predict strategic actions. Due to these limitations, we cannot reasonably expect to learn (or estimate) user functions in a gaming setting nor generalize results to broader scenarios. In real-life applications, the player utilities are not known a priori; therefore, the developed methods should have some way to account for data-driven learning techniques. In our past work, we have explored utility learning and incentive design as a coupled problem both in theory~\citep{konstantakopoulos2016inverse,konstantakopoulos2017robust, jia2018poisoning} and in practice~\citep{ratliff2014social,konstantakopoulos2016smart,konstantakopoulos2017leveraging} under a Nash equilibrium model. Our utility learning approaches are presented in a platform-based design flow for smart buildings~\citep{jia2018design,biscuit}. The broader purpose of this paper is to present a general learning framework that leverages game theoretic concepts for learning models of occupant decision making in a competitive setting and under a discrete set of actions.

Contemporary building energy management techniques employ a variety of algorithms in order to improve performance and sustainability. Many of these approaches leverage ideas from topics such as optimization theory and machine learning. Our goal was to improve building energy efficiency by introducing a gamification system that engages users in the process of energy management and integrates seamlessly through the use of a human-centric cyber-physical framework. There exists a considerable amount of previous work demonstrating the success of control and automation in the improvement of building energy efficiency~\cite{nagy2015building,ascione2017mpc}. Some other notable techniques implement concepts such as incentive design and adaptive pricing~\cite{Mathieu2012,he2018practical}. Modern control theory has been a critical source of inspiration for several approaches that employ ideas like model predictive and distributed control and have demonstrated encouraging results in applications like HVAC. Unfortunately, these control approaches lack the ability to consider the individual preferences of occupants, which highlights a significant advantage of human-centric scenarios over contemporary methods. While machine learning approaches are capable of generating optimal control designs, they fail to adjust to occupant preferences and the associated implications of these preferences to the control of building systems. The heterogeneity of user preferences in regard to building utilities is considerable and necessitates a system that can adequately account for differences from occupant to occupant.

Clearly, the presence of occupants greatly complicates the determination of an efficient building management system. With this in mind, focus has shifted toward modeling occupant behavior within the system in an effort to incorporate their preferences. To accomplish this task, the building and its occupants are represented as a multi-agent system targeting occupant comfort~\cite{nagy2015building}. First, occupants and managers are allowed to express their building preferences, and these preferences are used to generate an initial control policy. An iteration on this control policy is created by using a rule engine that attempts to find compromises between preferences. Some drawbacks of this control design are immediately apparent. There should be some form of communication to the manager about occupant preferences. In addition, there is no incentive for submission of true user preferences and no system for considering occupant feedback. Other related topics in the same vein focus on grid integration~\cite{samad:2016aa}, while still others consider approaches for policy recommendations and dynamic pricing systems \cite{Mathieu2012}.

As alluded to previously, the key to our approach is the implementation of a social game among users in a non-cooperative setting. Similar methods that employ \textit{social games} have been applied to transportation systems with the goal of improving flow~\citep{merugu:2009aa,pluntke2013insinc}. Entrepreneurial ventures have also sought to implement solutions of their own to the problem of managing building energy efficiency~\footnote{\tt \url{https://comfyapp.com}}~\footnote{\tt \url{https://coolchoices.com/how-it-works/improve}}~\footnote{\tt \url{https://www.keewi-inc.com/index.php}}. Finally, it has been shown that societal network games are useful in a \textit{smart city} context for improving energy efficiency and human awareness~\citep{de2014social}. The critical motivation behind the social game context is to create friendly competition between occupants. In turn, this competition will encourage occupants to individually consider their own energy usage and, hopefully, seek to improve it. This same gamification technique has also been used as a way to educate the public about energy usage~\citep{salvador:2012aa,orland:2014aa}. Additionally, it has been cleverly implemented in a system that presents feedback about overall energy consumption to occupants~\citep{simon:2012aa}. Another notable case of a gamification methodology was used to engage individuals in Demand Response (DR) schemes~\citep{li:2014aa}. In this application, each of the users were represented as an utility maximizer within the model of a nash equilibrium where occupants gain incentive for reduction in consumption during DR events. In contrast to approaches that target user devices with known usage patterns~\citep{li:2014aa}, our approach focuses on personal room utilities, such as lighting, without initial usage information, which simulates scenarios of complete ignorance to occupant behaviors. For our method, we utilize past user observations to learn the utility functions of individual occupants by way of several novel algorithms. Using this approach, we can generate excellent prediction of expected occupant actions. Our unique social game methodology simultaneously learns occupant preferences while also opening avenues for feedback. This feedback is generated through individual surveys that provide opportunities to influence occupant behavior with adaptive incentive. With this technique, we are capable of accommodating occupant behavior in the automation of building energy usage by learning occupant preferences and applying a variety of novel algorithms. Furthermore, the learned preferences can be adjusted through incentive mechanisms to enact improved energy usage.

A series of experimental trials were conducted to generate real-world data, which was then used as the main source of data for our approach. This differentiates our work from a large portion of other works in the same field that use simulations in lieu of experimental methods. In many cases, participants exhibit a tendency to revert to previously inefficient behavior after the conclusion of a game. Our approach combats this effect by implementing incentive design that can adapt to the behavior and preferences of occupants progressively, which ensures that continuous participant engagement.  

We use conventional machine learning algorithms as well as deep learning based algorithms to benchmark the forecast of energy resource usage by occupants (or their utility function). We use the graphical lasso algorithm as a powerful tool to understand the latent conditional dependence between variables~\citep{mainbook}. This in turn provides insights into how different features interplay among each other. Historically, Graphical Lasso has been used in various fields of science, ranging from study of how individual elements of the cell interact with each other~\citep{glasso_biology} and to the broad area of computer vision for scene labelling~\cite{cvpr_scene}. A modified version of the original algorithm, named time-varying graphical lasso, has been used on financial and automotive data~\citep{HallacPBL17}. However, the novelties of graphical lasso has not been well utilized in the area of energy cyber-physical systems. We use Granger causality\citep{granger1980testing} to explain the causal relationship between the features in energy usage behavior of agents in social game. It has been widely used in the energy domain in applications such as deducing the causal relationship between economic growth and energy consumption~\citep{chiou2008economic}.

With this social game framework, a building manager will be capable of considering the individual preferences of the occupants within the scope of the building's energy consumption. This social game system could potentially offer an unprecedented amount of control for managers without sacrificing occupant comfort and independence.


\section{Smart Building Social Game: Implementation \& Human Decision-Making}\label{social_game}


In this section, we introduce our proposed social game concept as a gamification application implemented at Nanyang Technological University (NTU) residential housing apartments, along with the software architecture design for the deployed Internet of Things (IoT) sensors. In addition to the implementation of this gamification application, we abstract the agent decision-making processes in a game theoretic framework and introduce the discrete choice theory that we draw upon for forecasting agent actions with high accuracy.

\subsection{Description of the Social Game Experiment}

Our experimental environment is comprised of residential housing single room apartments on the Nanyang Technological University (NTU) campus. The residential housing single room apartments on the NTU campus were divided into four blocks, each of which had two floors. In this space, there were a total of seventy-two occupants who were eligible to participate in the social game. Participation in our social game platform was voluntary. We ran the experiment in both the Fall 2017 and Spring 2018 semesters. In the Fall 2017 version, we included ceiling light, desk light, and ceiling fan resources in the graphical user interface for the social game, while in the Spring 2018 version we included all of the potential resources that were available.

We designed a social game web portal so that all single room dorm occupants could freely view their daily room's energy resource usage with a convenient interface. In each dorm room, we installed two Internet of Things (IoT) sensors \footnote{\textit{IoT sensor tag}: \url{http://www.ti.com/ww/en/wireless_connectivity/sensortag/index.html}}, one close to the desk light and another near the ceiling fan. With the deployment of IoT sensors, dorm occupants can monitor in real-time their room's lighting system (desk and ceiling light usage) and HVAC (ceiling fan and A/C usage) with a refresh interval of up to one second. 

Dorm occupants were rewarded with points based on how energy efficient their daily usage is in comparison to their past usage before the social game was deployed. The past usage data that serves as our baseline is gathered by monitoring occupant energy usage for approximately one month before the introduction of the game for each semester. Using this prior data, we calculated a weekday and weekend baseline for each of the occupant's resources. We accumulate data separately for weekdays and weekends so as to maintain fairness for occupants who have alternative schedules of occupancy (e.g. those who tend to stay at their dorm room over the weekends versus weekdays). We employ a lottery mechanism consisting of several gift cards awarded on a bi-weekly basis to incentivize occupants, i.e. occupants with more points are more likely to win the lottery. Earned points for each resource is given by the following equation:

\begin{equation}
  \hat{p}^{d}_{i}(b_i, u_{i})= s_{i} \frac{b_i - u^{d}_{i}}{b_i}
  \label{eq:points_earned}
\end{equation}

where $\hat{p}^{d}_{i}$ is the points earned at day $d$ for room's resource $i$ which corresponds to ceiling light, desk light, ceiling fan, and A/C. Also, $b_i$ is the baseline calculated for each resource $i$, $u^{d}_{i}$ is the usage of the resource at day $d$, and $s_{i}$ is a points booster for inflating the points as a process of framing \citep{tversky1981framing}. This process of framing can greatly impact a user's participation, and it is routinely used in rewards programs for credit cards among many other point-based programs used in industry. In addition, we rewarded dorm occupants for the percentage of savings (equation \ref{eq:points_earned}) because it is important to motivate all of the participants to optimize their usage independent of the total amount of energy consumed in their normal schedule. However, over-consumption resulted in negative points.


In Figure~\ref{fig:game_utilization_design_A}, we present how our graphical user interface is capable of reporting to occupants the real-time status (on/off) of their devices, their accumulated daily usage, time left for achieving daily baseline, and the percentage of allowed baseline being used, by hovering above the utilization bars. In order to boost participation, we introduced a randomly appearing coin close to the utilization bars with the purpose of incentivizing occupants to log in to the web portal and view their usage. The coin was designed to have a psychological impact on the occupants, i.e. to motivate occupants towards viewing their resource usage and understanding their impact to energy consumption by getting exact usage feedback in real-time. Based on this game principle, we gave occupants points when they clicked on the coin, which could increase both their perceived and actual chances of winning rewards.


\subsection{Internet of Things (IoT) System Architecture}

\begin{figure}[!ht]
    \subfloat[Graphical user interface (GUI)]{%
    \includegraphics[width=0.46\textwidth]{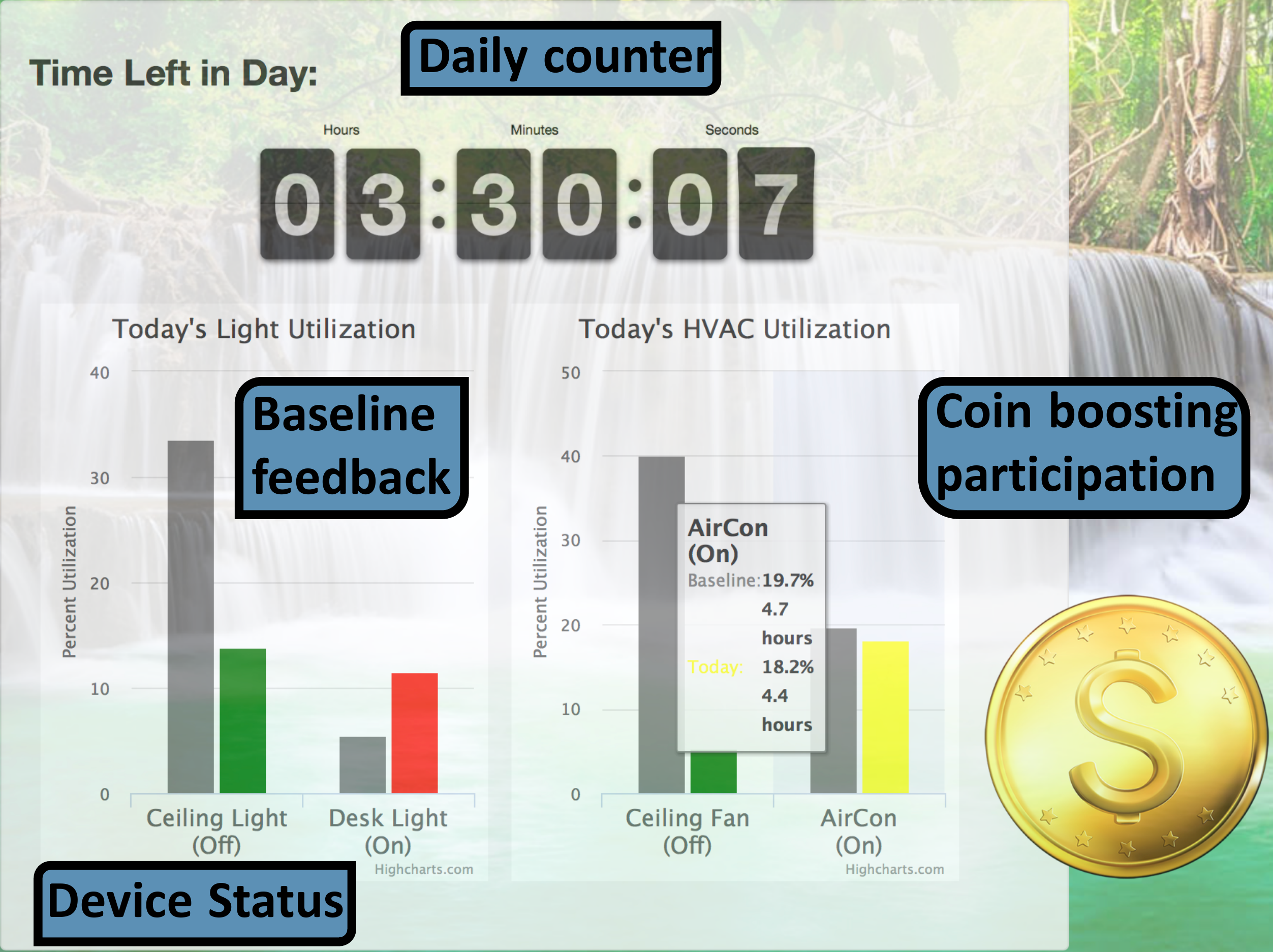}\label{fig:game_utilization_design_A}
    }
    \hfill
    \subfloat[Social game dataflow architecture design]{%
    \includegraphics[width=0.45\textwidth]{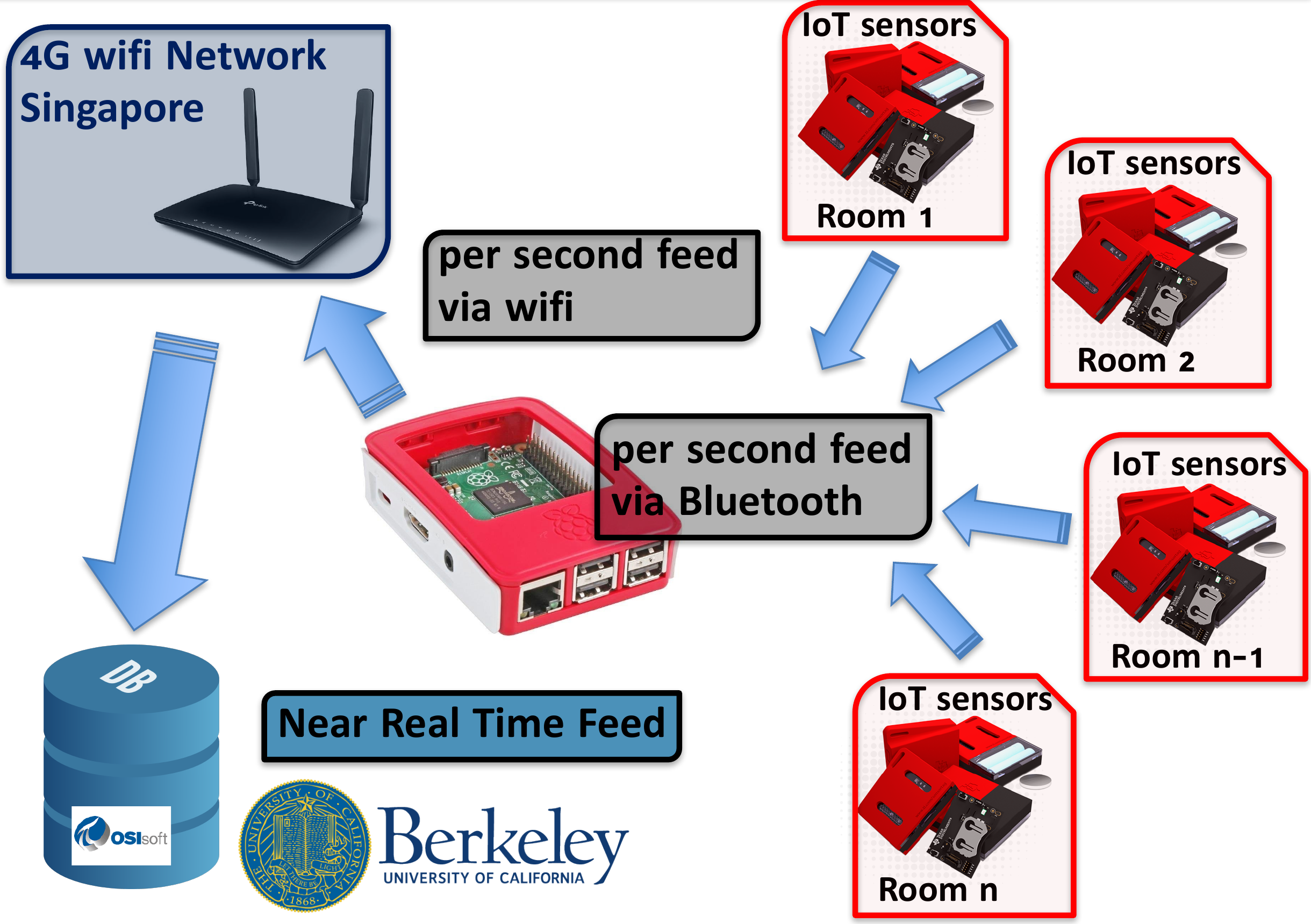}\label{fig:game_utilization_design_B}
    }
\caption{Graphical user interface (GUI) and dataflow design for social game}
\label{fig:game_utilization_design}
\end{figure}

We enabled the design and implementation of a large-scale networked social game through the utilization of cutting-edge Internet of Things (IoT) sensors. In total, we deployed one hundred and forty-four sensors in single dorm rooms. These were part of a hardware and software infrastructure that achieved near real-time monitoring of various metrics of resource usage in each room, e.g. lighting and A/C. Moreover, our system was capable of saving occupant actions in the web portal. Weather data was gathered from an externally-installed local weather monitoring station at per-second resolution. The actual design and dataflow is depicted in Figure~\ref{fig:game_utilization_design_B}.

Utilizing the data gathered from each dorm room, we leveraged several indoor metrics like indoor illuminance, humidity, temperature, and vibrations for the ceiling fan sensor. Having performed various tests during Summer 2017 within the actual unoccupied dorm rooms, we derived simple thresholds indicating if a resource is in use or not. For instance, the standard deviation of acceleration gathered from the ceiling fan mounted sensor is an easy way to determine whether the ceiling fan is in the on state. Additionally, by combining humidity and temperature values, we were able to reliably identify whether A/C is in use with limited false positives. Our calibrated detection thresholds were robust over daylight patterns, external humidity/temperature patterns, and measurement noise introduced by IoT sensors.

While we receive data from various dorm room sensors, our back-end processes update the status of the devices in near real-time in each occupant's account and update points based on their usage and point formula (equation \ref{eq:points_earned}). This functionality allows occupants to receive feedback, view their points balance, check rankings, and more. In order to allow participants to assess and visualize their energy efficient behavior, each user's background in the web portal changed based on their ranking and energy efficiency. We used background pictures of rain forest settings for encouraging more energy efficient behavior and images of desert scenes to for occupants with limited energy savings.


\subsection{Agent Decision-Making Model}\label{game_model}

Discrete choice theory is greatly celebrated in the literature as a means of data-driven analysis of human decision-making. Under a discrete choice model, the possible outcome of an agent can be predicted from a given choice set using a variety of available features describing either external parameters or characteristics of the agent. We use a discrete choice model as a core abstraction for describing occupant actions related to their dorm room resources.  

Consider an agent $i$ and the decision-making choice set which is mutually exclusive and exhaustive. The decision-making choice set is indexed by the set $\mc{I}=\{\mc{J}^1,\ldots, \mc{J}^S \}$. Decision maker $i$ chooses between $S$ alternative choices and would earn a \textit{representative utility} $f_i$ for $i \in \mc{I}$. Each decision among decision-making choice set leads agents to get the highest possible utility, $f_i > f_j$ for all $i,j \in \mc{I}$. In our setting, an agent has an utility which depends on a number of features $x_z$ for $z=1, \ldots, N$. However, there are several unobserved features of the representative utility which should be treated as random variables. Hence, we define a \textit{random utility} decision-making model for each agent given by

\begin{equation}
  \hat{f}_i(x)=g_i(\beta_{i},x) + \epsilon_{i}
  \label{eq:discrete_utility}
\end{equation}

where $\epsilon_{i}$ is the unobserved random component of the agent's utility, $g_i(\beta_{i},x)$ is a nonlinear generalization of agent $i$'s utility function,
and where

\begin{equation}
x=(x_1, \ldots, x_{i-1}, x_{i+1}, \ldots, x_N)\in \mb{R}^{N} 
\label{eq:feature}
\end{equation} 


is the collective $n$ features explaining an agent's decision process. The choice of nonlinear mapping $g_i$ and $x$ abstracts the agent's decision; it could represent, e.g., how much of a particular resource they choose to use and when an agent optimizes its usage over a specific resource. Discrete choice models in their classical representation~\cite{hensher2018discrete} are given by a linear mapping $g_i(\beta_{i},x) = \beta_{i}^{T}x$ in which $\epsilon_{i}$ is part of independently and identically distributed random variables modeled using a Gumbel distribution.




\subsection{Game Formulation}

To model the outcome of the strategic interactions of agents in the deployed social game, we use a \emph{sequential non-cooperative discrete game} concept. To introduce our generalized decision-making model for each agent (equation ~\ref{eq:discrete_utility}), a sequential non-cooperative discrete game is given by,

\begin{definition}

Each agent $i$ has a set $\mc{F}_i = {f_{i}^1,\ldots, f_{i}^{N}}$ of $N$ \textit{random utilities}. Each random utility $j$ has a convex decision-making choice set $\mc{I}_{j}=\{\mc{J}^{1}_{j},\ldots, \mc{J}^{S}_{j} \}$. Given a collection of $n$ features~\eqref{eq:feature} comprising the decision process and the temporal parameter $T$, agent $i$ faces the following optimization problem for their \textit{aggregated random utilities}:

\begin{equation}
  \max\{\sum_{i=1}^{N}f^{T}_i(x) | f_i \in \mc{F}_i\}.
  \label{eq:opt-seq}
\end{equation}
\end{definition}


Like a sequential equilibrium concept, we simulate the game defined by the estimated random utility functions per resource to demonstrate the actual decision-making process of each individual dorm occupant. Agents in the game independently co-optimize their aggregated random utilities (equation ~\ref{eq:opt-seq}) given a collection of $n$ features (equation ~\ref{eq:feature}) at each time instance $T$. A general incentive design mechanism (equation ~\ref{eq:points_earned}) (seen in the building level of the gamification framework in Figure~\ref{fig:gamification_abstraction_A}) motivates their potential actions across various given decision-making choice sets. The above definition extends the definition of a discrete choice model~\citep{hensher2018discrete} to sequential games in which agents co-optimize several discrete, usually mutually exclusive, choices.



\subsection{Social Game Data Set} \label{social_game_data_set}

As a final step, we aggregate occupant data in per-minute resolution. We have several per-minute features like time stamp, resource status, accumulated resource usage in minutes per day, resource baseline, gathered points (both from game and surveys), occupant ranking over time, and number of occupant visits to the web portal. In addition to these features, we add several external weather metrics like humidity, temperature, and solar radiation.

After gathering a high-dimensional data set with all of the available features, we propose a pooling and picking scheme to enlarge the feature space and then apply a Minimum Redundancy and Maximum Relevance (mRMR)~\citep{peng2005feature} feature selection procedure to identify useful features for our predictive algorithms. We pool additional features from a subset of the already derived features by leveraging domain knowledge. Specifically, we consider two different feature types: dummy features (using one-hot encoding) and resource features. Dummy features represent intangible variables relating to weekly or seasonal events such as weekends in the former case and holidays in the latter. Resource features include deterministic data sets gathered by our instrumentation such as daily percentage of resource usage. We open source the dataset after proper benchmarking.
  


\section{Benchmarking and Explainability Analysis}
\label{machine_learning}


In this section, we explore the utility learning problem using Deep Learning methods that serve to improve forecasting accuracy. From a broader perspective, our goal is to demonstrate how the proposed learning scheme fits into the overall gamification abstraction in Figure~\ref{fig:gamification_abstraction_and_block_diagram}. We will show that our utility learning methods lead to accurate energy usage forecasts, which in turn can be integrated in demand response programs (seen in the provider/retailer level in Figure~\ref{fig:gamification_abstraction_A}). This goal motivates why we are interested in learning more than a simple predictive model for agents, but rather an utility-based forecasting framework that accounts for individual preferences, dynamic changes in agent behavior, and heterogeneous actions.


\subsection{Benchmarking using Conventional Machine Learning Framework}


In section~\ref{game_model}, we introduced an extension to discrete choice models for sequential decision-making over a set of different choices. More concretely, we examine the utility learning problem using a novel pipeline including a variety of statistical learning methods and models that improve estimation and prediction accuracy for our proposed sequential discrete choice model. Furthermore, well-trained classification models serve as an excellent benchmark for our proposed Deep Learning models.

\subsubsection{Random Utility Estimation Pipeline}

We start by describing the basic components of our proposed random utility estimation pipeline using and data gathered from the game played between agents and pooled features. Let us now introduce the benchmarking pipeline formulation as it serves as the basis for the random utility estimation.




After gathering streaming data in our MySQL data-base (as described in Section~\ref{social_game}), we pool several candidate features and expand our feature space. Next, a large set of proposed high-dimensional candidate features is constructed. Using this feature set, we adopt a greedy feature selection algorithm called Minimum Redundancy Maximum Relevance (mRMR)~\citep{peng2005feature}. The mRMR greedy heuristic algorithm utilizes mutual information as the metric of goodness for a candidate feature set. Given the large number of pooled candidate features, mRMR feature selection is a useful method for finding a subset of features that are relevant for the prediction of occupant resource usage. The mRMR feature selection algorithm is applied to batched data from the game period either in the Fall or Spring version of the Social Game. From the total number of available features, we decided to keep nearly half of them.

After getting a number of important features as a result of the mRMR greedy algorithm, we apply a simple data pre-processing step with mean subtraction across each individual feature. Mean subtraction centers the cloud of data around the origin along every dimension. On top of mean subtraction, we normalize the data dimensions by dividing each dimension by its standard deviation in order to achieve nearly identical scale in the data dimensions. However, the training phase of the random utility estimation pipeline has one potentially significant challenge, which is the fact that data in almost every resource is heavily imbalanced (e.g. the number of resources with off samples is on the order of 10-20 times more than those with on samples). This is expected considering occupants daily patterns of resource usage in buildings, but it poses a risk of having potentially poorly trained random utility estimation models.

For optimizing around highly imbalanced data sets, we adapt the Synthetic Minority Over-Sampling (SMOTE)~\cite{chawla2002smote} technique for providing balanced data sets for each resource and for boosting prediction (e.g. classification) accuracy. SMOTE over-samples a data set used in a classification problem by considering $k$ nearest neighbors of the minority class given one current data point of this class. The SMOTE algorithm can be initialized by leveraging a pre-processing phase with Support Vector Machines as a grouping.


After the SMOTE step, we train several classifiers as a final step for the random utility estimation pipeline. We propose a base model of logistic regression. In an effort to improve this discrete choice model, we include penalized logistic regression (regLR) with $l_1$ norm protocol (Lasso) for the model training optimization procedure, among other classical classification machine learning algorithms. We perform a randomized grid search for optimizing classifiers using the Area Under the Curve (AUC) metric~\citep{majnik2013roc}, aiming to co-optimize TPR (sensitivity) and FPR (1-specificity). 



We use the Area Under the Receiver Operating Characteristic (ROC) Curve as our performance metric. ROC curves describe the predictive behavior of a binary classifier by plotting the probability of true positive rate (i.e. correct classification of samples as positive) over false positive rate (i.e. the probability of falsely classifying samples as positive). For training the proposed machine learning algorithms, we used k-fold cross validation combined with the AUC metric to randomly split the data into training and validation sets in order to quantify the performance of each proposed machine learning model in the training phase. Each machine learning algorithm used in our benchmark pipeline and their respective hyper-parameters are described in more depth in ~\citet{konstantakopoulos2018deep}.

\subsection{Leveraging Deep Learning for Sequential Decision-Making}

In this section, we formulate a novel deep learning framework for random utility estimation that allows us to drastically reduce our forecasting error by increasing model capacity and by structuring intelligent deep sequential classifiers. The architecture for deep networks is adaptive to proposed sequential non-cooperative discrete game models and achieves a tremendous increase in forecasting accuracy. Hence, deep networks achieve an end-to-end training for modeling agents random utility (equation ~\ref{eq:discrete_utility}) with exceptional accuracy. Due to ease of access to big data and the rapid development of adaptive artificial intelligence techniques, energy optimization and the implementation of smart cities has become a popular research trend in the energy domain. Researchers have deployed Deep Learning and Reinforcement Learning techniques in the field of energy prediction~\citep{mocanu2016deep,fan2017short} and intelligent building construction~\citep{2016intelligent}.



In our framework of random utility learning in a non-cooperative game setting, deep networks work as powerful models that can generalize our core model by increasing capacity and by working towards an intelligent machine learning model for predicting agent behavior. 

\subsubsection{Deep Neural Networks for Decision-Making}

\begin{figure}[!ht]
    \subfloat[Architecture of proposed deep neural network]{%
    \includegraphics[width=0.285\textwidth]{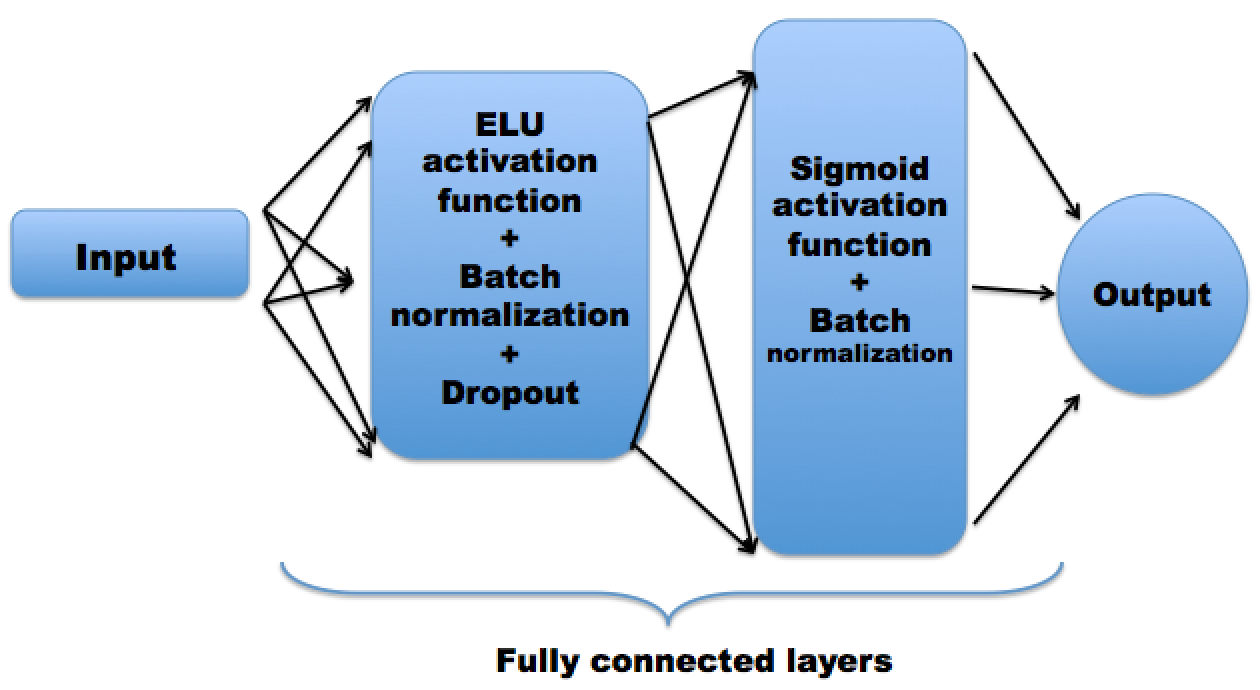}\label{fig:DNN_DRNN_A}
    }
    \hfill
    \subfloat[Architecture of deep bi- directional recurrent neural networks]{%
    \includegraphics[width=0.315\textwidth]{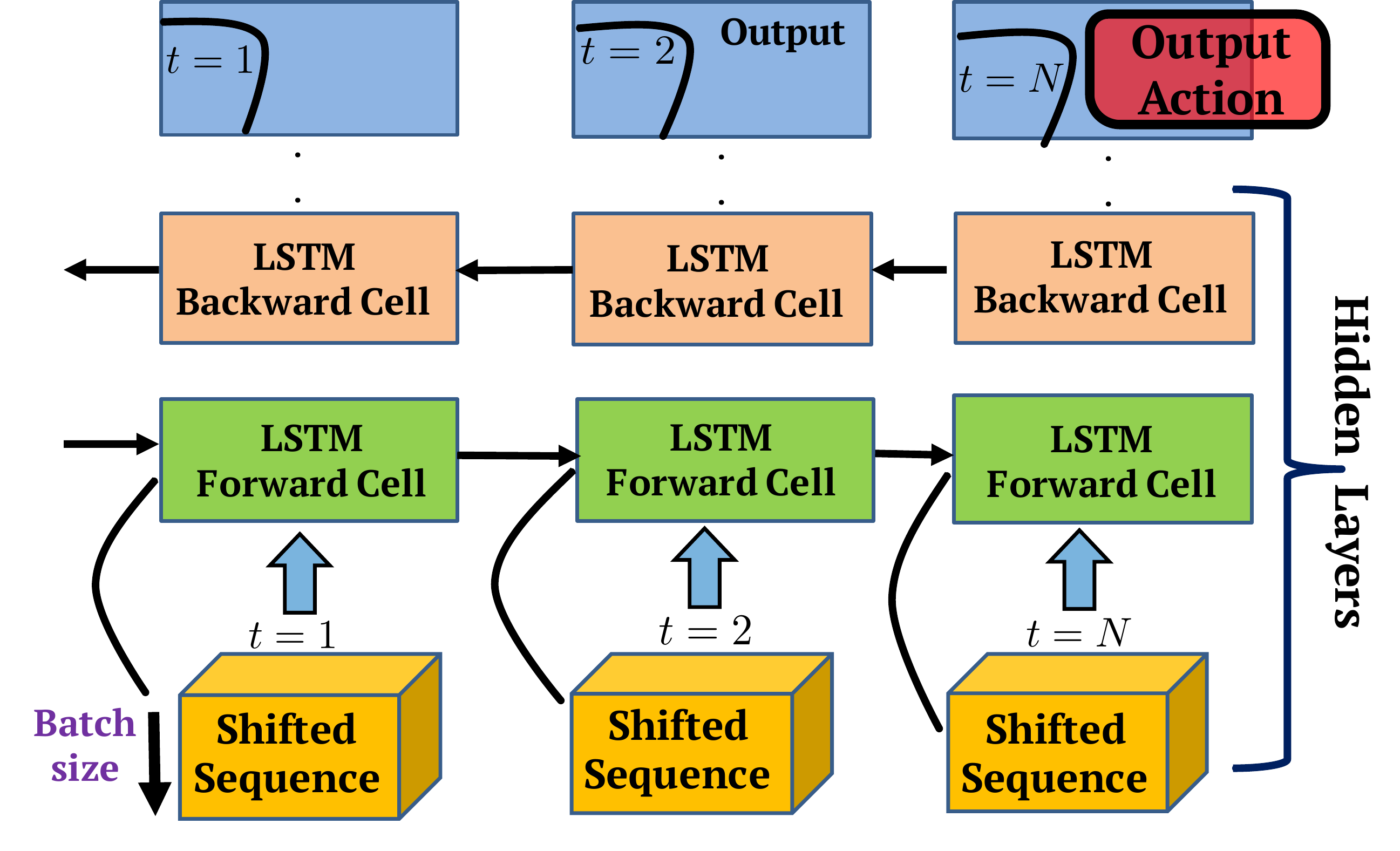}\label{fig:DNN_DRNN_B}
    }
    \hfill
    \subfloat[Conventional and recurrent based auto-encoder]{%
    \includegraphics[width=0.365\textwidth]{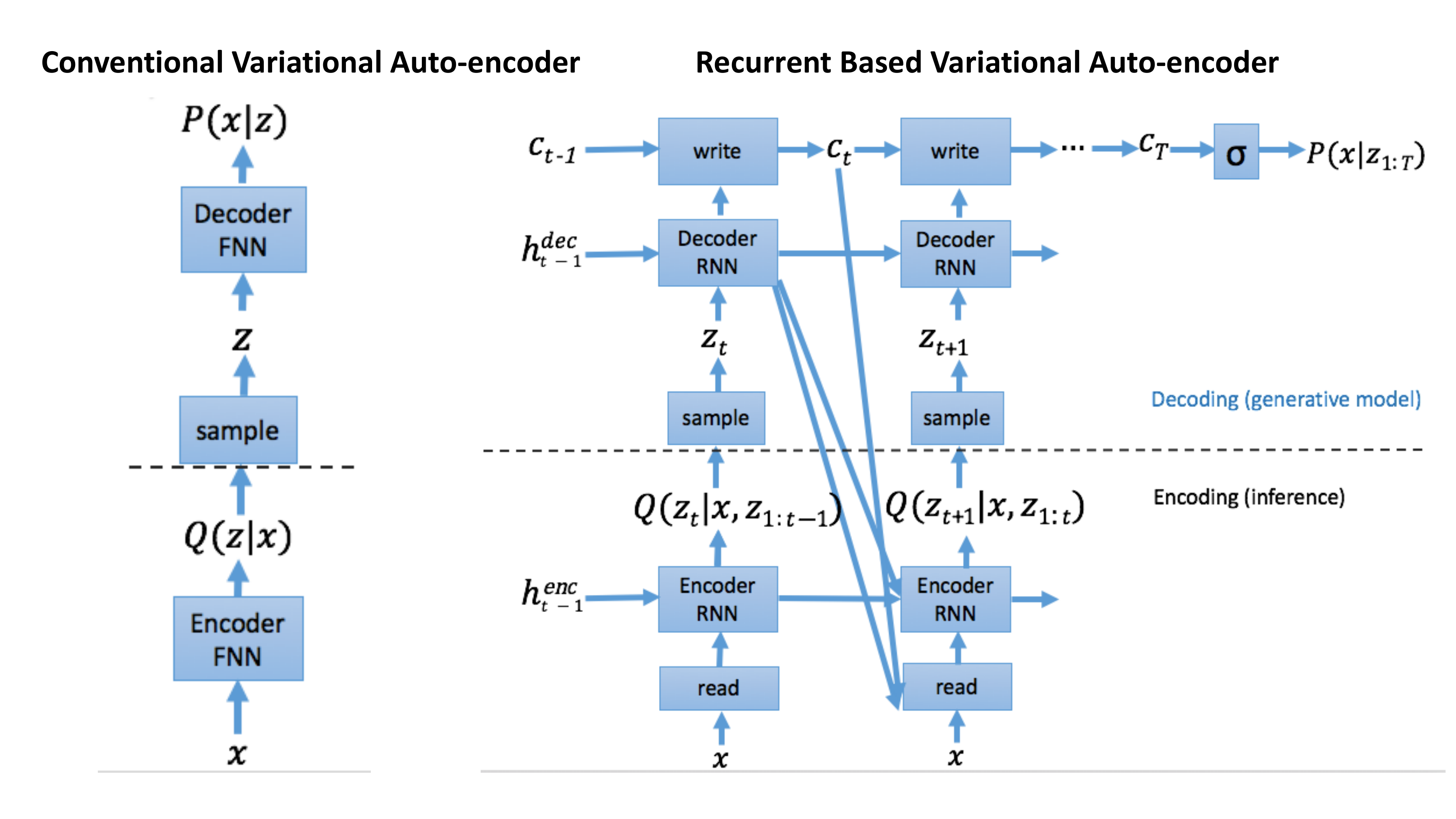}\label{fig:DNN_DRNN_C}
    }
    \hfill
\caption{Proposed deep neural networks, deep bi-directional recurrent neural network, and auto-encoders}
\label{fig:DNN_DRNN}
\end{figure}

Deep neural network techniques have drawn ever-increasing research interests ever since Deep Learning in the context of rapid learning algorithms was proposed in 2006~\cite{hinton2006fast}. Our approach using neural networks has the inherent capacity to overcome deficiencies of the classical methods that are dependent on the limited series of features located in the training data set (e.g. such as the features resulting from mRMR in our setting). A deep neural net can be seen as a typical network in which the input flows from the input layer to the output layer through a number of hidden layers. An illustration of the deep neural network for our random utility learning is depicted in Figure~\ref{fig:DNN_DRNN_A}. 

Our proposed deep neural network model for random utility learning includes exponential linear units (ELUs) ~\citep{goodfellow2016deep} at each hidden layer. The usage of exponential linear units ~\citep{goodfellow2016deep} normally adds additional hyper-parameter in the search space as a trade-off for significant increase in fitting accuracy due to enormous decrements of "dead" units, a classical problem of rectified linear unit (ReLU) implementations ~\citep{clevert2015fast}. The output layer is modeled using sigmoid units for classifying agents discrete choices. The proposed model is optimized by minimizing the cross-entropy cost function using stochastic gradient descent combined with a nesterov optimization scheme. The initialization of the weights utilizes He normalization ~\citep{he2015delving}, which gives increased performance and better training results. Unlike a random initialization, He initialization avoids local minimum points and makes convergence significantly faster. Batch Normalization ~\citep{ioffe2015batch} has also been adopted in our deep neural network framework to improve the training efficiency and to address the vanishing/exploding gradient problems in the training of deep neural networks. By using batch normalization, we avoid drastic changes in the distribution of each layers inputs during training while the deep network parameters of the previous layers keep changing. Knowing that adding more capacity in our deep neural network model will potentially lead to over-fitting, we apply dropout technique ~\citep{srivastava2014dropout} as a regularization step. The dropout technique involves the following procedure in the training phase (both in forward and backward graph learning traversal steps): each neuron, excluding the output neurons, has a probability to be totally ignored. The probability to ignore a neuron is another hyper-parameter of the algorithm and normally gets values between 50\% - 70\%.

\subsubsection{Deep Bi-directional Recurrent Neural Networks for Sequential Decision-Making}

One of the basic drawbacks of both benchmark random utility learning models as well as the proposed deep neural networks is that they have strong assumptions for the data generation process. An important challenge for efficient learning of sequential decision-making models is the actual modeling of the dependence of future actions of an agent with the present and previous actions. In particular, an agent naturally tries to co-optimize around a set of discrete choices and gains the higher utility (equation ~\ref{eq:opt-seq}). Both benchmark models and deep neural networks adopt the assumption of independent and identically distributed data points. One way to model the underlying time series dependencies is through efficient feature engineering and by potentially using a novel feature selection algorithm. In Section~\ref{social_game_data_set}, we use domain knowledge along with a pooling \& picking method to create a feature set that can accurately predict agents' behavior. However, this step helps sparingly in the presence of time series dependencies and cannot generalize. 

Leveraging the latest Deep Learning models, like recurrent neural networks, we address the issue of time dependence by looking at temporal dependencies within the data. Recurrent neural networks have the capability to allow information to persist, even over long periods, by simply inserting loops that point to them. Lately, recurrent neural networks have been implemented with huge success in energy and automotive sectors. Specifically, recurrent neural networks can be applied to energy related fields such as wind energy conversion systems~\citep{zhao2016novel,shi2018direct} and solar radiation prediction~\citep{gensler2016deep,al2017nonlinear}. In the automotive sector, recurrent neural networks are used for anticipating driver activity~\citep{jain2016recurrent} in addition to autonomous driving testing~\citep{tian2018deeptest}. 


As we see in the architecture of a deep bi-directional recurrent neural network in Figure~\ref{fig:DNN_DRNN_B}, information passes from one time step of the network to the next. The information of the network passes to successor nodes. In the case of a bi-directional recurrent neural network, information can also flow in the opposite direction to the predecessor. In a simple implementation, however, recurrent neural networks tend to either vanish or become incapable of modeling long-term dependencies. In our proposed novel sequential utility learning model, we enable an end-to-end training using Long Short Term Memory cells (LSTM). Mainly, LSTM includes several gates that decide how long and short-term relations should be modeled. The overall output of the LSTM cell is a combination of sub-gates describing the dependencies~\citep{goodfellow2016deep}. 






Our deep bi-directional architecture is described in Figure~\ref{fig:DNN_DRNN_B}. Given an agent's actions, we define a time step $N$ (sliding window of actions), which is a hyper-parameter and models time series dependencies in agent actions. Each training instance in the network is a tensor with the following dimensions: mini-batch size (using a multiple of time step $N$), target sequence, and all given features.


Our deep network is designed to leverage sequential data and build several layers and time steps of hidden memory cells (LSTMs). Moreover, we propagate the unrolled deep network both forward and backward (bi-directional recurrent neural network) for modeling the exact series time-lagged features for agents actions. For the proposed deep bi-directional recurrent neural network, we use three hidden layers. To perform classification for the agent actions, we add a fully connected layer. The fully connected layer is in turn connected to the output of the ending time-step propagating network, which is finally followed by a soft-max layer, which is what actually performs the classification task. Similar to the deep neural network model, we optimize by minimizing the cross-entropy cost function using stochastic gradient descent combined with a Nesterov optimization scheme. Additionally, we employ an exponentially decaying learning rate as our learning rate schedule. Again, the initialization of the weights utilizes He normalization~\citep{he2015delving}, which gives increased performance and better training results. To avoid over-fitting with the enormous capacity of the proposed deep network, we apply dropout as the core regularization step.

\subsubsection{Deep Learning for Generative Sequential Decision Models}

On top of the existing data set resulting from our experiment, we can create even larger data sets based on the existing ones. The idea of bootstrapping~\citep{Hastie09} is widely applied both in statistics and machine learning in many applications to help with the creation of new data sets that mimic the original data. However, bootstrapping is not a scalable solution. As data sets become larger and larger, the computational complexity restricts the capabilities of the system. In our approach, we propose the use of deep variational auto-encoders~\citep{kingma2014auto} as an approach to create a nonlinear manifold (encoder) that can be used as a generative model. Variational auto-encoders formalize the necessary generative model in the framework of probabilistic graphical models by maximizing a lower bound on the log-likelihood of the given high-dimensional data. Furthermore, variational auto-encoders can fit large high-dimensional data sets (like our social game dataset) and train a deep model to generate data resembling the original data set. In a sense, generative models automate the natural features of a data set and then provide a scalable way to reproduce known data. This capability can be employed either in the utility learning framework for boosting estimation or as a general way to create simulations mimicking occupant behavior/preferences in software like EnergyPlus \footnote{https://\url{energyplus.net}}.

Using such a Deep Learning model, we can generate data by simply sampling a latent vector from the latent space of the auto-encoder and decoding via the decoder component. In Figure~\ref{fig:DNN_DRNN_C}, we provide the overall architecture behind training a variational auto-encoder. We use two hidden layers in encoder and decoder while tying parameters between them. Also, the latent space is modeled using a standard gaussian distribution. By using this architecture of deep auto-encoder, however, we limit the generative model in applications in which the data process has a natural time-series dependence. Hence, we proposed the implementation of a recurrent variational auto-encoder~\citep{gregor2015draw}. In its architecture shown in Figure~\ref{fig:DNN_DRNN_C}, the proposed recurrent variational auto-encoder allows time-series modeling for progressive refinement and spatial attention in the shifted tensor inputs. Using progressive refinement, the deep network simply breaks up the joint distribution over and over again in several steps resulting in a chain of latent variables. This gives the capability to sequentially output the time-series data rather than compute them in a single shot. Moreover, a recurrent variational auto-encoder can potentially improve the generative process over the spatial domain. By adding time series in the model as tensors with shifted data points, we can reduce the burden of complexity by implementing improvements over small regions of the tensor input at a time instance (spatial attention). 

With these mechanisms, we achieve reduction of the complexity burden that an auto-encoder has to overcome. As a result, using a recurrent based variational auto-encoder allows for more generative capabilities that can handle larger, more complex distributions such as those in the given social game time series data. Our models were tested in several sets of data from individual occupants and were capable of randomly generating new data with significant similarities to the training data. This result provides a powerful tool for generating new data on top of the existing data and provides more flexibility in the application of the data in several real scenario mechanisms like demand response. 

\subsection{Explainability Analysis using Graphical Lasso}
With the growth of artificial intelligence, there has been a pressing need for models and datasets to have explainable nature. For our proposed social game data, we ensure it has explainability, i.e. the data correctly encodes the human decision making towards energy usage in competetive environments. We divide the players into three classes in a supervised way taking the ranks of the users as the label. Let the classes be represented by $C^{High}$, $C^{Medium}$ and $C^{Low}$, where the superscripts signify the energy efficiency behavior of each class. The behavior of a player in a particular class represents the characteristic behavior of players, e.g. $C^{High}$ represents the behavior of players showcasing high energy efficiency. We then study the feature correlations in different classes using graphical lasso algorithm. We formulate a framework that allows us to understand users decision making model. Specifically, we adopt graphical lasso algorithm ~\cite{friedman2008sparse,mainbook} to study the way in which features in a energy game-theoretic framework interplay among each other.

Let the features representing the social game data be denoted by the collection $Y = (Y_1, Y_2, \cdots, Y_H)$. From a graphical perspective, $Y$ can be associated with the vertex set $V =  \{1, 2, \cdots, H\}$ of some underlying graph. The structure of the graph is utilized to derive inferences about the relationship between the features. We use the graphical lasso algorithm ~\cite{mainbook} to realize the underlying graph structure, under the assumption that the distribution of the random variables is Gaussian.

Consider the random variable $Y_h$ at $h$ $\in$ $V$. We use the Neighbourhood-Based Likelihood for graphical representation of multivariate Gaussian random variables. Let the edge set of the graph be given by $\it{E}$ $\subset$ $V\times V$. The neighbourhood set of $Y_h$ is defined by 
\begin{align}
    \mathcal{N}(h) = \{k \in V | (k,h) \in \textit{E}\}
\end{align}
and the collection of all other random variables be represented by:
\begin{align}
    {Y}_{V\backslash\{h\}} = \{Y_{k}, k\in (V-\{h\})\}
\end{align}
For undirected graphical models, node for $Y_h$ is conditionally independent of nodes not directly connected to it given $Y_{\mathcal{N}(h)}$, i.e.
\begin{align}
    (Y_h|{Y}_{V\backslash\{h\}}) \sim (Y_h|{Y}_{\mathcal{N}(h)})
\end{align}
The problem of constructing the inherent graph out of observations is nothing but finding the edge set for every node. This problem becomes predicting the value of $Y_h$ given $Y_{\mathcal{N}(h)}$, or equivalently, predicting the value of $Y_h$ given $Y_{\backslash\{h\}}$by the conditional independence property. The conditional distribution of $Y_h$ given $Y_{\backslash\{h\}}$ is also Gaussian, so the best predictor for $Y_h$ can be written as:
\begin{align}
    {Y}_{h} = {Y}_{V\backslash h}^T.{\beta}^{h} + {W}_{V\backslash h}
\end{align}
where $W_{V\backslash h}$ is zero-mean gaussian prediction error. The $\beta^{h}$ terms dictate the edge set for node $h$ in the graph. We use $l_1$-regularized likelihood methods for getting a sparse $\beta^{h}$. Let the total number of data samples available be \textit{N}. The optimization problem is formulated as: corresponding to each vertex $h = 1, 2, \cdots ,H$, solve the following lasso problem:
\begin{align}
    \hat{{\beta}^{h}} \in \underset{\beta^h \in {\mathbb{R}}^{H-1}}{\rm argmin} \bigg\{{\frac{1}{2N}\sum_{j=1}^{N}(y_{jh}-{y^T_{j,{V\backslash h}}}}\beta^{h})^2 + \lambda{\|\beta^h\|}_{1}\bigg\}
\end{align} 
The implementation of Graphical Lasso algorithm is summarized in Appendix.\\
\section{Experimental Results}
\label{results}


We now present the results of the proposed utility learning method applied to high-dimensional data collected from the social game experiment in the Fall 2017 and Spring 2018 semesters. As previously described, our data set consists of the per-minute high-dimensional data of occupant energy usage across several resources in their rooms. We evaluate the performance of utility learning under two characteristic scenarios. The first scenario involves having full information from the installed IoT sensors for performing "step-ahead" predictions. In this scenario, IoT sensors continuously feed information from the previous actions of the occupants. For the second scenario, referred to as "sensor-free", we stop taking the IoT sensor readings into account in each room. In the second instance, the aggregated past features of the occupants are missing. For this case, we have a model in which we only use features that we can acquire from external weather conditions (e.g. from a locally installed weather station), information about occupant engagement with the web portal, and seasonal dummy variables. All of these features are much easier to acquire without needing to keep the highly accurate but expensive IoT devices. The broader purpose of our proposed gamification approach is the development of exceptional forecasting models representing dynamic occupant behavior. As illustrated in Figure~\ref{fig:gamification_abstraction_A}, the building level energy usage prediction is fed to upper level components of the smart grid at the provider or microgrid level. In a real application scenario, the proposed building level modeling opens new avenues for demand response programs, which can incorporate real-time predictions of building occupant energy patterns. The proposed game theoretic models and iterative incentive design mechanisms are powerful in the sense that they can simultaneously be used to predict but also to incentivize desirable human behavior. Adaptive incentive design motivates building occupant energy efficiency through gamification platforms while accurately predicting their energy usage in order to feed it back to the higher provider levels of our framework.

From our experiment, we present estimation results for the data set in both Fall and Spring versions of the experiment for two characteristic occupants. Both occupants have ample data for all of the relevant resources being considered. We used the first four game periods for the training of our models:


 \begin{itemize}
     \setlength\itemsep{0.05em}
     \item Fall: Sep. $12^{\text{th}}$, 2017 - Nov. $19^{\text{th}}$, 2017 ($n=100,800$); Nov. $20^{\text{th}}$, 2017 - Dec. $03^{\text{rd}}$, 2017 ($n=20,160$).
     \item Spring: Feb. $19^{\text{th}}$, 2018 - Apr. $22^{\text{nd}}$, 2018 ($n=90,720$); Apr. $23^{\text{rd}}$, 2018 - May $06^{\text{th}}$, 2018 ($n=20,160$).
 \end{itemize}


Before we trained our benchmark classifiers, we applied the mRMR algorithm to the total data set (data from all occupants) in the training period. This accounts for almost 4 million distinct data points in the Fall semester data set and 2.5 million distinct data points in the Spring semester data set. Applying mRMR results in several top features in both the Fall and Spring semester data sets. Interestingly, mRMR included several external features in the top relevant feature candidates. In particular, the presence of external humidity as an important feature for the ceiling fan is a good demonstration of the mRMR algorithm's capability to extract salient features. Moreover, features like survey points illustrate that some occupants co-optimized their resource usage while also trying to view their point balance, usage, and ranking in the game (by visiting the web portal).



\subsection{Forecasting via Conventional Machine Learning \& Deep Learning Frameworks}

We have dual objectives in our leveraging of conventional machine learning and Deep Learning frameworks. Our first objective is to achieve highly accurate forecasts of building occupant resource usage. Providers and retailers at the higher levels of Figure~\ref{fig:gamification_abstraction_A} can integrate energy usage forecasts in demand response programs. Our second objective is to improve building energy efficiency by creating an adaptive model that learns how user preferences change over time and thus generate the appropriate incentives to ensure active participation. Furthermore, the learned preferences can be adjusted through incentive mechanisms~\cite{ratliff2014social} to enact improved energy efficiency (seen in the building level of Figure~\ref{fig:gamification_abstraction_A}).

For learning optimal random utility models in the benchmark setting, we use the top twenty-five resulting features from the mRMR algorithm along with a pre-processing step of SMOTE with SVM initialization. Using SMOTE, we boost the accuracy of benchmark models due to the fact that our data set was heavily imbalanced. We achieve decent accuracy using well-trained conventional machine learning models. Area Under the Receiver Operating Characteristic Curve is our forecasting performance metric. All of the classifiers achieve decent AUC scores in both the Fall and Spring semester results, as shown in Table~\ref{tab:AUC_all}. In the \textquotedblleft sensor-free\textquotedblright results, we have a significant drop in the achieved accuracy, but this is expected given that the IoT feed is decoupled. However, even in "sensor-free" examples we are able to predict occupant behavior using less representative features and having excluded the IoT sensors. 

For the deep neural networks, we used training data resulting from the applied SMOTE step. We used two hidden layers of the feed-forward neural network, with 50\% dropout and stochastic gradient descent method leveraging nesterov's Momentum to accelerate convergence. 20\% of the training data was used as a validation data set for hyper-parameter tuning. 


To further exploit the continuity of the sequential decision-making model, we experiment on the bi-directional deep recurrent neural network. We used a time sliding window---time step of two hours (120 distinct data points). We processed the data without being pre-processed from the SMOTE algorithm as we wanted to retain the underlying sequence of actions of the occupants (temporal dependencies of the data). We used three hidden layers with 60\% dropout rate, and we applied an exponentially decaying learning rate (simulated annealing). In the training of bi-directional recurrent neural networks, we applied the principle of early stopping using a validation data set over the AUC metric. For our deep bi-directional networks, thirty-five epochs were optimal to be trained. As in deep neural networks, 20\% of the training data was used as validation set for hyper-parameter tuning. 

\begin{figure}[t]
    \subfloat[AUC Scores using Fall semester data]{%
    \includegraphics[width=0.48\textwidth]{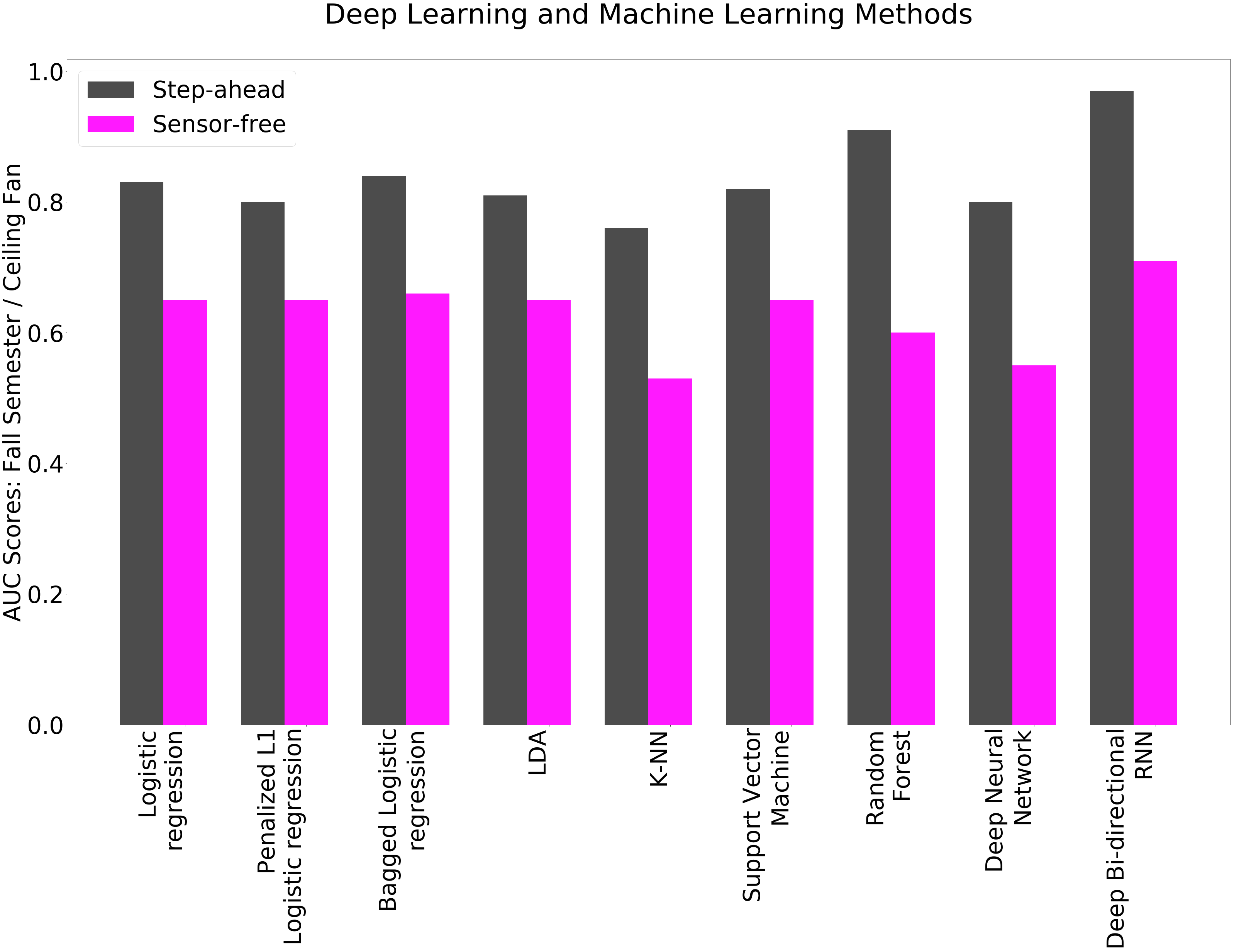}
    }
    \hfill
    \subfloat[AUC Scores using Spring semester data]{%
    \includegraphics[width=0.48\textwidth]{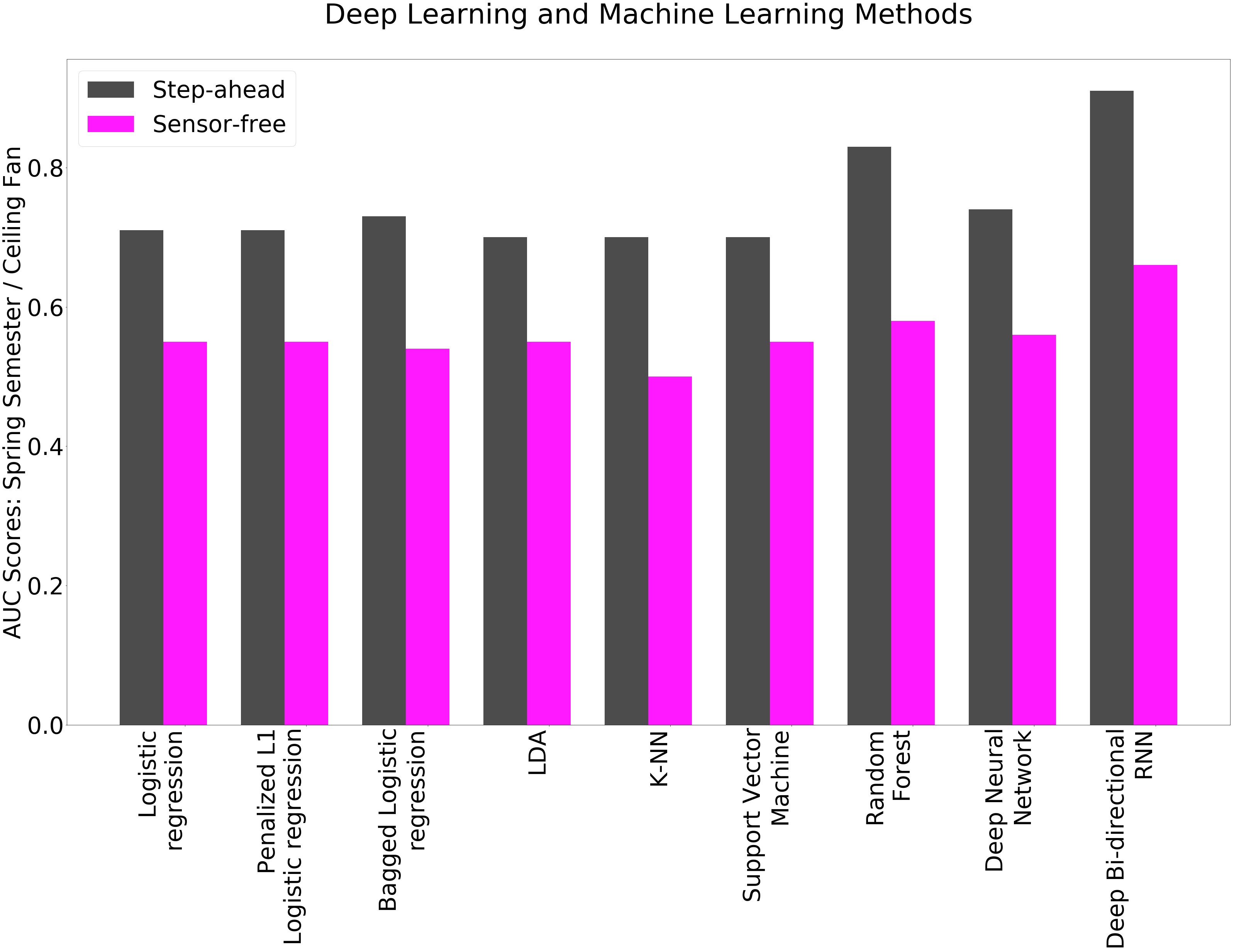}
    }
\caption{Forecasting accuracy ("step-ahead" / "sensor-free" predictions) for ceiling fan usage (on/off).}
\label{fig:AUC_predictions}
\end{figure}

\begin{table}[ht!]
\begin{center}
\setlength\arrayrulewidth{0.5pt}
\caption{AUC scores for \textquotedblleft step-ahead\textquotedblright/\textquotedblleft sensor-free\textquotedblright predictions.}
\label{tab:AUC_all}
\scalebox{1}{
\resizebox{\textwidth}{!}{\begin{tabular}[!h]{ |c|c|c|c|c|c|c|c|}
\rowcolor{Gray}
\hline 
  & \multicolumn{3}{|c|}{Fall Semester} & \multicolumn{4}{|c|}{Spring Semester}\\
 \hline
 \textquotedblleft step-ahead\textquotedblright / \textquotedblleft sensor-free\textquotedblright & Ceiling Fan & Ceiling Light & Desk Light & Ceiling Fan & A/C & Ceiling Light & Desk Light\\
 \hline
Logistic regression &  0.83 / 0.65 &  0.78 / 0.61 &  0.78 / 0.68 & 0.71 / 0.55  & 0.76 / 0.73 & 0.75 / 0.55 & 0.76 / 0.50\\
 \hline
Penalized $l1$ Logistic regression &  0.80 / 0.65 &  0.77 / 0.56 & 0.78 / 0.64 & 0.71 / 0.55 & 0.76 / 0.70 & 0.75 / 0.55  & 0.76 / 0.50\\
\hline
 Bagged Logistic regression &  0.84 / 0.66 &  0.80 / 0.59 &  0.79 / 0.68 & 0.73 / 0.54 & 0.73 / 0.73 & 0.76 / 0.54 & 0.79 / 0.51\\
 \hline
LDA &   0.81 / 0.65 &  0.78 / 0.58 &  0.74 / 0.68 & 0.70 / 0.55 & 0.73 / 0.73 & 0.75 / 0.55 & 0.70 / 0.51\\ 
\hline
K-NN &   0.76 / 0.53 &  0.77 / 0.56 &  0.74 / 0.55 &  0.70 / 0.50 & 0.76 / 0.57 & 0.68 / 0.54 & 0.73 / 0.57\\ 
\hline
Support Vector Machine &  0.82 / 0.65 & 0.78 / 0.60 &  0.76 / 0.68 &  0.70 / 0.55  & 0.75 / 0.73 & 0.75 / 0.55 & 0.70 / 0.50\\ 
\hline
Random Forest &  0.91 / 0.60 &  0.78 / 0.59 &  0.98 / 0.63 & 0.83 / 0.58 & 0.83 / 0.65 & \textbf{0.99} / 0.54 & 0.96 / 0.50\\ 
\hline
 Deep Neural Network &  0.80 / 0.55 &  0.76 / 0.60 &  0.78 / 0.64 & 0.74 / 0.56 & 0.78 / 0.68 & 0.77 / 0.54 & 0.84 / 0.50\\
\hline
Deep Bi-directional RNN &  \textbf{0.97} / \textbf{0.71} & \textbf{0.85} / \textbf{0.66} & \textbf{0.99} / \textbf{0.76} & \textbf{0.91} / \textbf{0.66} & \textbf{0.89} / \textbf{0.80} & \textbf{0.99} / \textbf{0.64} & \textbf{0.99} / \textbf{0.62}\\ 
\hline
\end{tabular}}}
\end{center}
\end{table}

To evaluate the effectiveness of our proposed deep learning framework, we present the AUC scores of a representative example for comparison. From the results, it is clear that deep RNN outperforms the majority of alternative algorithms. One important remark is that deep RNN exceeds even the random forest algorithm, which is considered a top-performing, robust classification model. Deep NN also achieved better performance in some examples over the Random Forest classifier, but this is not a general case. Figure~\ref{fig:AUC_predictions} introduces bar charts representing AUC scores for ceiling fan usage (on/off). Prediction results are divided into AUC scores for the two scenarios discussed previously, \textquotedblleft step-ahead\textquotedblright and \textquotedblleft sensor-free\textquotedblright. Upon examination of these results, it is clear that deep RNN outperforms all other deep learning and machine learning models. Figure~\ref{fig:AUC_predictions} demonstrates that deep bi-directional RNN based models achieve accuracy almost equal to one. For more results, please refer to  \citet{konstantakopoulos2018deep}.

\subsection{Generative Modeling via Sequential Deep Auto-Encoders}
\begin{table}[t]
\begin{center}
\setlength\arrayrulewidth{0.5pt}
\caption{Feature comparison between proposed generative models (auto-encoders) using DTW score.}
\label{tab:autoencoder_res}
\scalebox{0.8}{
\begin{tabular}[!h]{|l|l|l|l|}
\rowcolor{Gray}
\hline
Time Series Feature  & Conventional Auto-encoder & RNN-based Auto-encoder & p-values \\
\hline
Ceiling Fan Status (On / Off)          &  1.5e+04 &  \textbf{1.2e+04} &  0.11 \\
\hline
Ceiling Light Status (On / Off)        &  1.6e+04 &  \textbf{2.2e+03} &  1.0 \\
\hline
Desk Light Status (On / Off)           &  6.7e+03 &  \textbf{0.0e+00} &  1.0 \\
\hline
Dorm Room Temperature                &  1.3e+05 &  \textbf{1.2e+05} &  0.0 \\
\hline
Dorm Room Humidity            &  4.8e+05 &  \textbf{3.7e+05}
 &  0.0 \\
\hline
External Temperature               &  1.0e+05 &  1.8e+05 &  0.0 \\
\hline
External Humidity          &  2.9e+05 &  4.3e+05 &  0.0 \\
\hline
\end{tabular}}
\end{center}
\end{table}

In Table~\ref{tab:autoencoder_res}, we present the results of two trained generative models using the full data set of a randomly selected occupant in the Fall semester. We trained both a conventional auto-encoder and a recurrent based auto-encoder. The resulting deep generative models can be used as a way to create simulations for mimicking building occupant behavior and preferences. This is an extra tool for quantifying variations in building occupant behavior as dynamic parts of a microgrid. Generative models are capable of adapting to variations of the external weather conditions, which in turn creates an interesting view of building occupant energy usage patterns aligned with external weather patterns.

In Table~\ref{tab:autoencoder_res}, we present results for several selected features from the interior of a dorm room and from external weather data. For evaluating artificially generated time-series from the proposed auto-encoders, we utilize dynamic time warping (DTW), which measures the similarity between two temporal sequences, the ground truth and the artificial data~\citep{rakthanmanon2012dtw}. In bold, we see that the recurrent based auto-encoder achieves a smaller DTW score in most of the features, leading to a generative model that isn't mimicking exactly or is deviating significantly from the original data set. 

In efforts to evaluate the statistical significance of the calculated DTW scores from the recurrent based auto-encoder, we used a permutation hypothesis test. In this approach, we permute original and generated time-series data, and we compute their DTW score looking for events that are more "extreme" than the one that is presented in Table~\ref{tab:autoencoder_res}. Interestingly, we have inside and outside weather based features (temperature and humidity) that have zero p-values, showing that the DTW scores using a recurrent based auto-encoder are statistically significant. For indoor device status features, however, p-values are large, indicating DTW score has high variability under the permutation test.

\subsection{Explainability Analysis}
\subsubsection{Feature correlation learning using graphical lasso}
We learn the feature correlations using graphical lasso algorithm in classes $C^{High}$, $C^{Medium}$ and $C^{Low}$ to obtain the knowledge about factors governing human decision making towards various (high/medium/low) energy efficient behaviors. We consider a representative player (selected as the player holding the median rank in the class) for each of the three classes to run graphical lasso and study the correlation between the features for that class. We group the features into different categories so as to study their influence on energy efficiency behaviors. Specifically, we consider \emph{Temporal} features like time of the day, academic schedules and weekday/weekends, \emph{External} features as outdoor temperature, humidity, rain rate etc. and \emph{Game Engagement} features like frequency of visits to game web portal. 

\begin{figure}[!ht]
    \subfloat[]{%
    \includegraphics[width=0.4\textwidth]{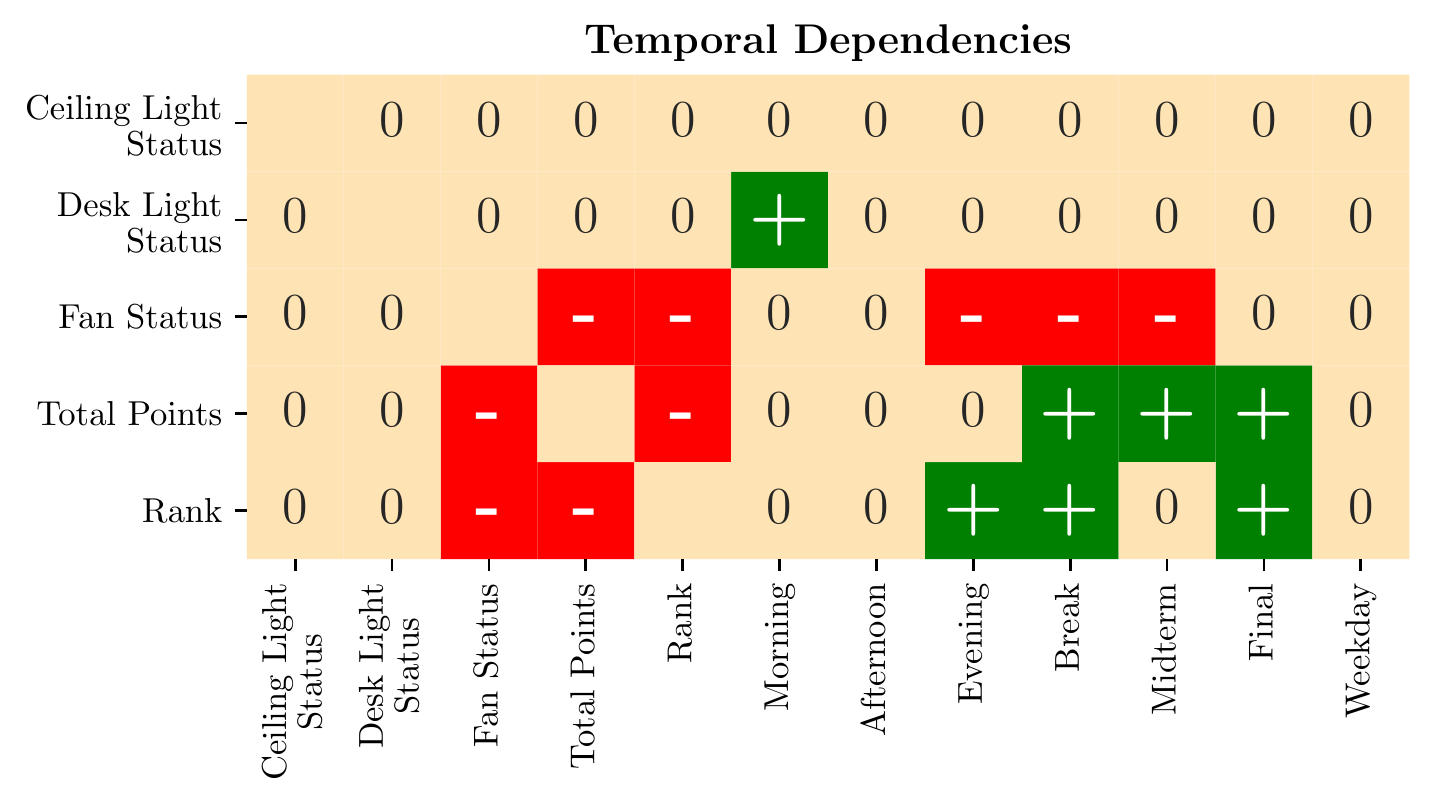}
    }
    \subfloat[]{%
    \includegraphics[width=0.33\textwidth]{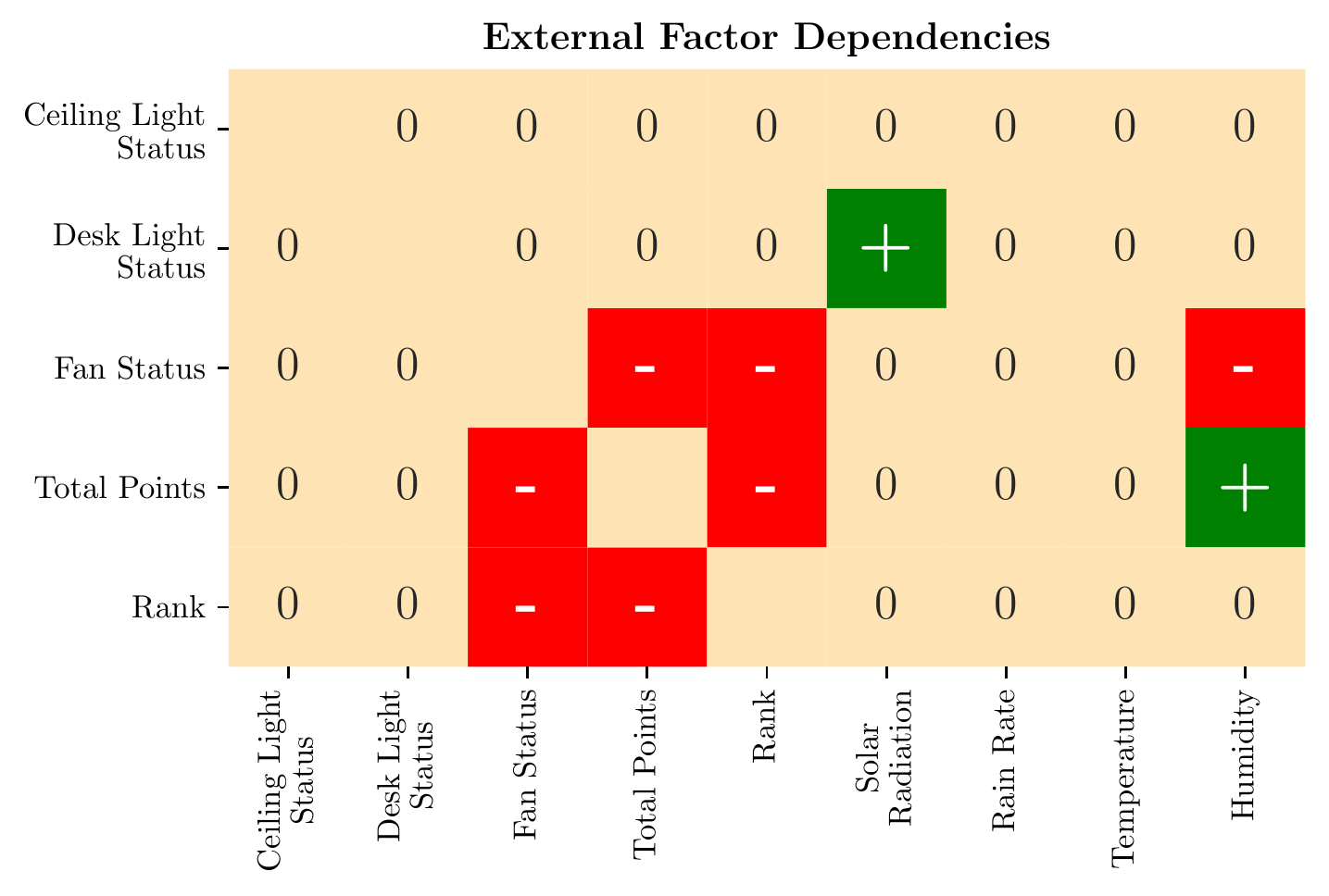}
    }
    \subfloat[]{%
    \includegraphics[width=0.25\textwidth]{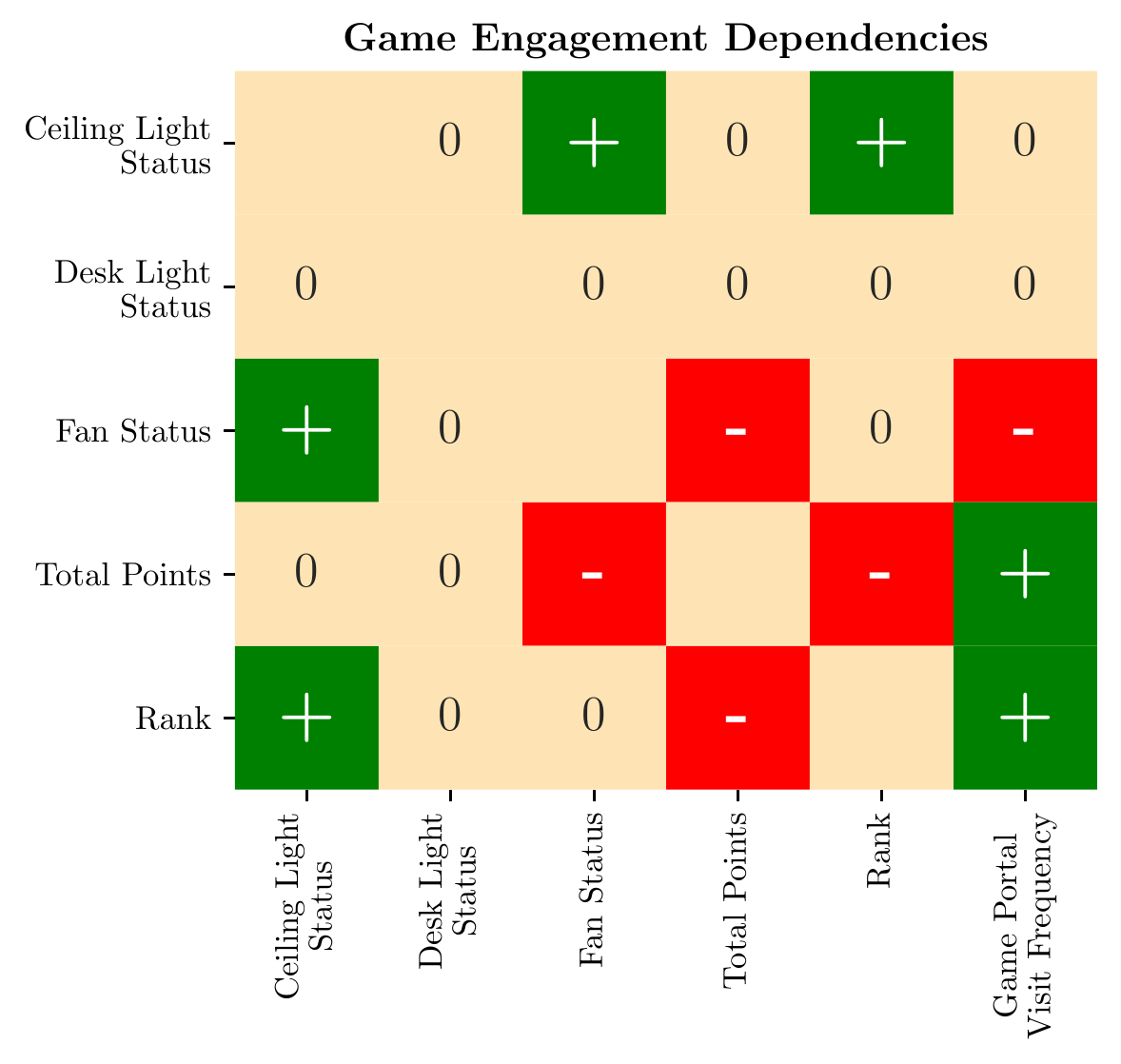}
    }
\caption{{\color{black}Feature correlations for a Low Energy Efficient Player ($\in$ $C^{Low}$)}}\label{fig_low}
\end{figure}

The feature correlations for a low energy efficient player is given in Fig~\ref{fig_low}. The player tries to use each resource independently which can be observed in Figure~\ref{fig_low}(a) with no correlation between the corresponding resource usage identifiers. There is a positive correlation between morning and desk light usage indicating heedless behavior towards energy savings. The absolute energy savings 
\newpage
\begin{figure}[t]
\centering
    { \centering                             
        \stackunder[5pt]\subfloat{\includegraphics[height=0.7cm]{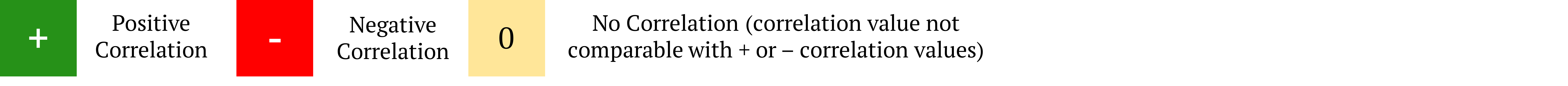}}
    }
\end{figure}
\begin{figure}[!ht]
    \subfloat[]{%
    \includegraphics[width=0.4\textwidth]{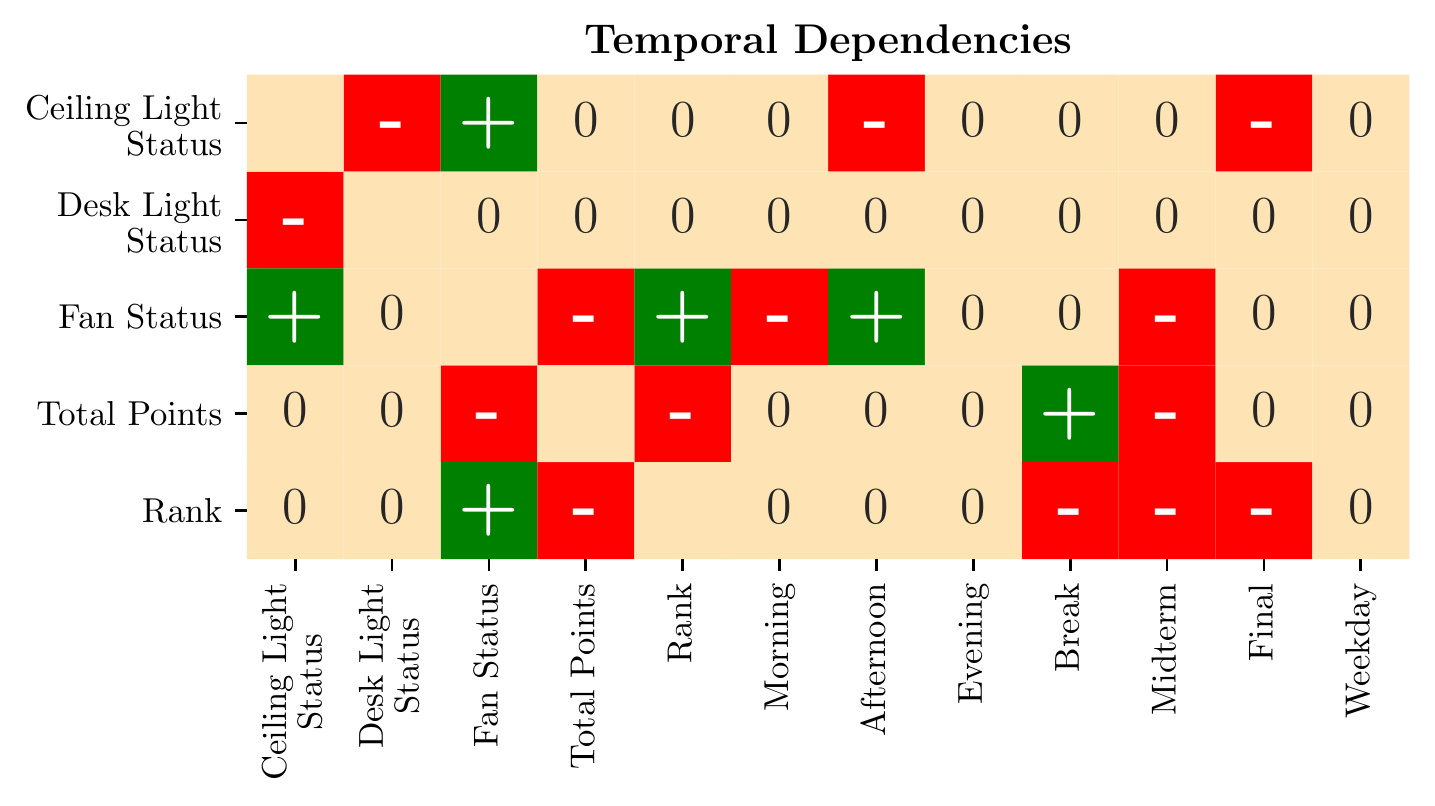}
    }
    \subfloat[]{%
    \includegraphics[width=0.33\textwidth]{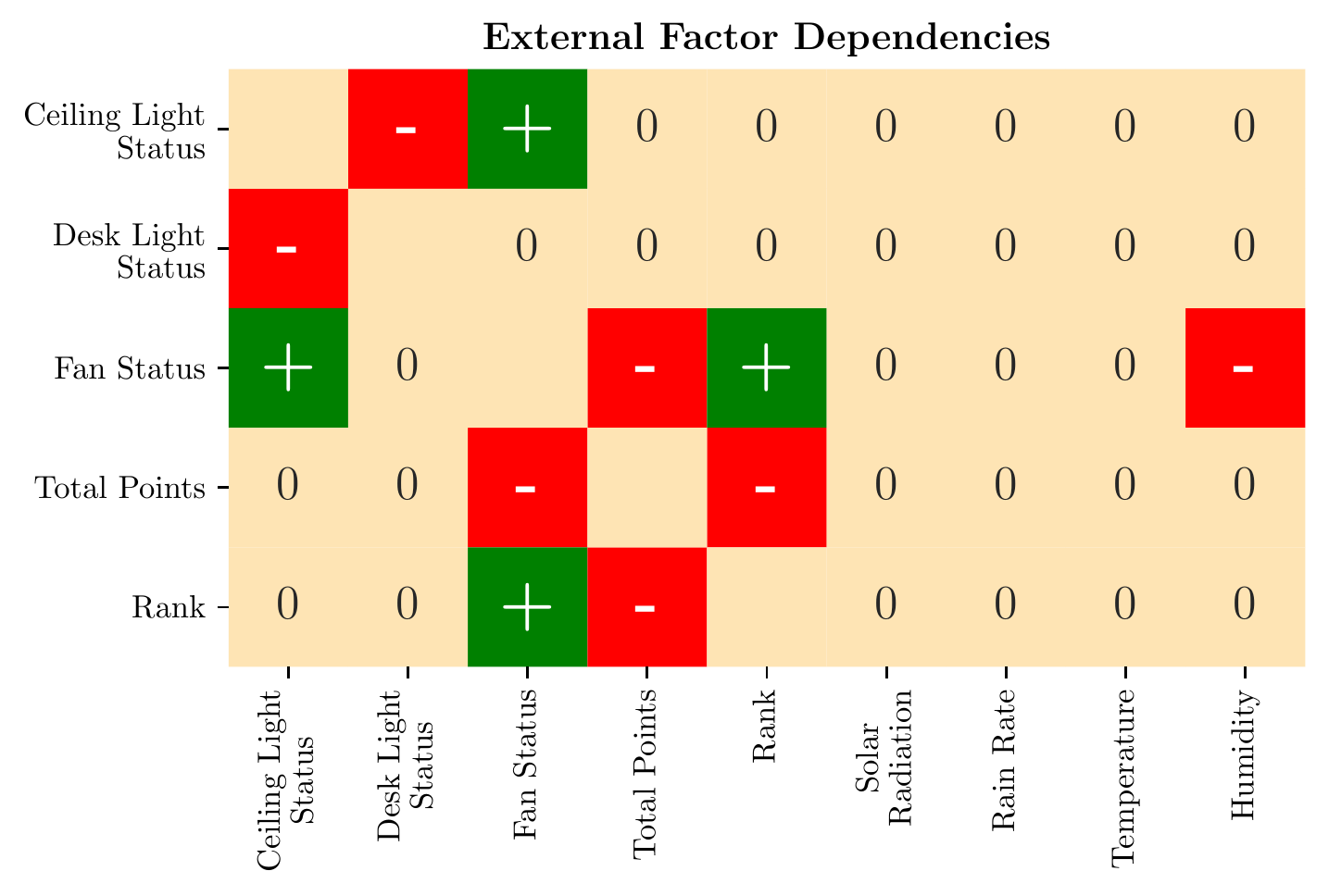}
    }
    \subfloat[]{%
    \includegraphics[width=0.25\textwidth]{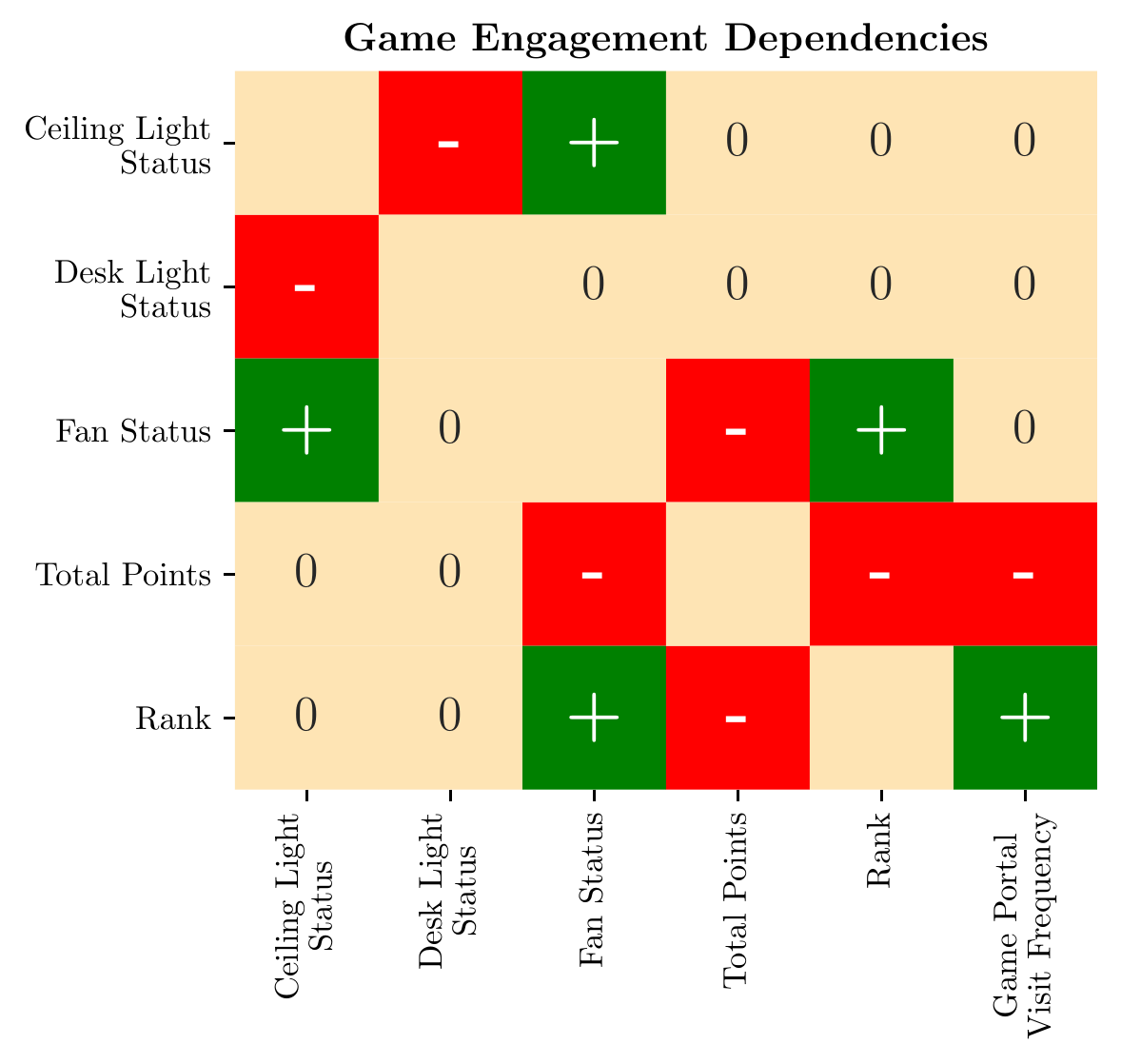}
    }
\caption{{\color{black}Feature correlations for a Medium Energy Efficient Player ($\in$ $C^{Medium}$)}}\label{fig_med}
\end{figure}
\begin{figure}[!ht]
    \subfloat[]{%
    \includegraphics[width=0.4\textwidth]{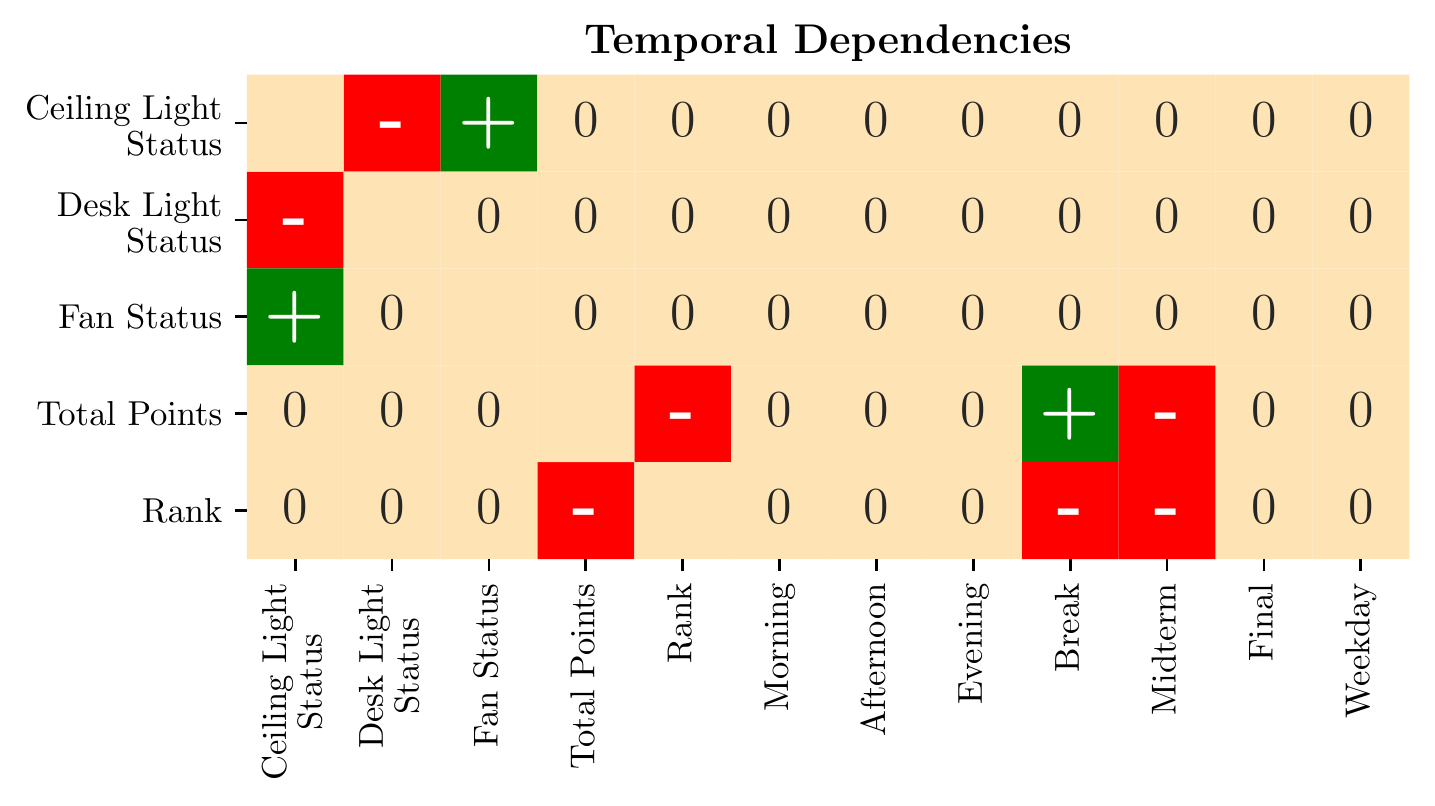}
    }
    \subfloat[]{%
    \includegraphics[width=0.33\textwidth]{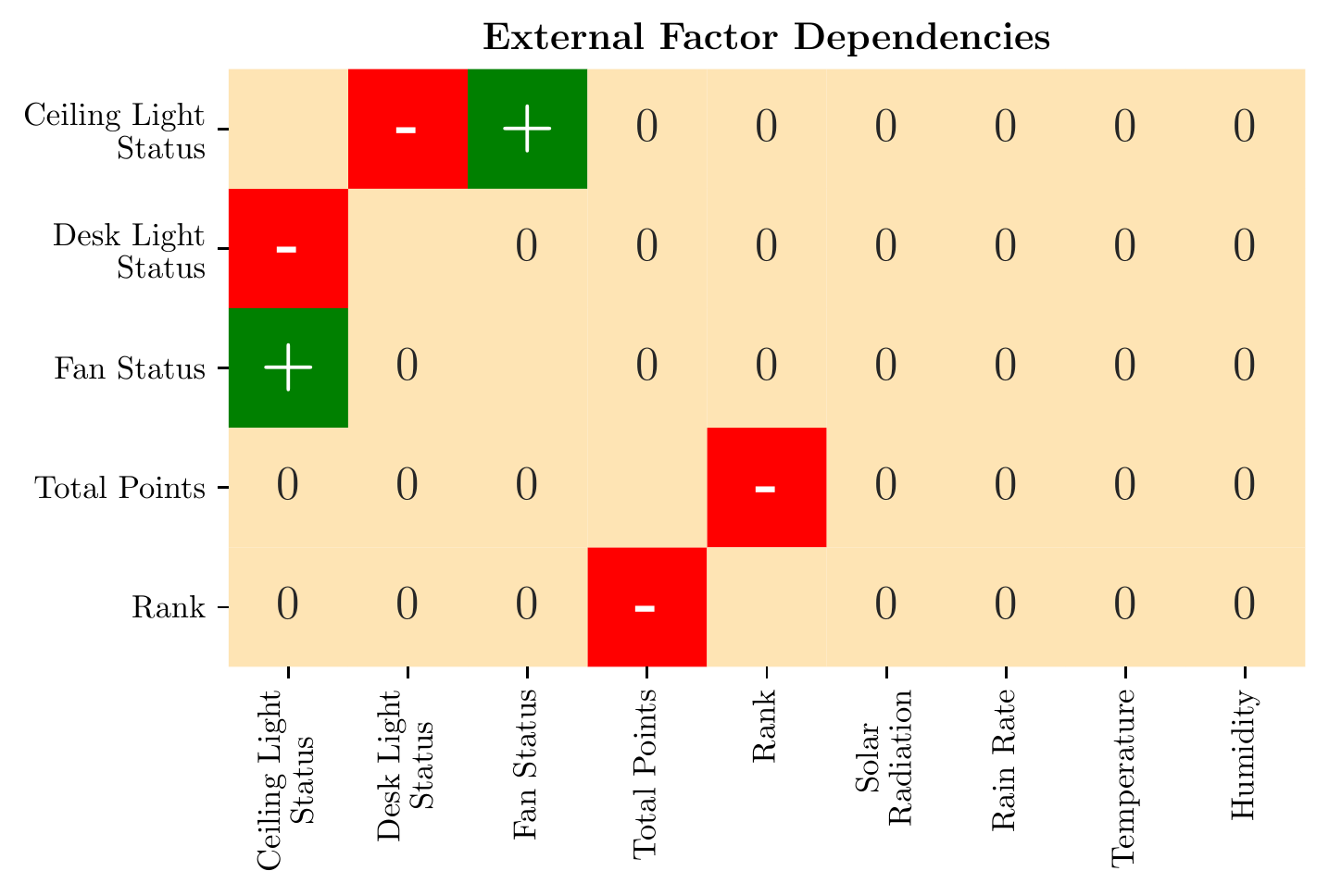}
    }
    \subfloat[]{%
    \includegraphics[width=0.25\textwidth]{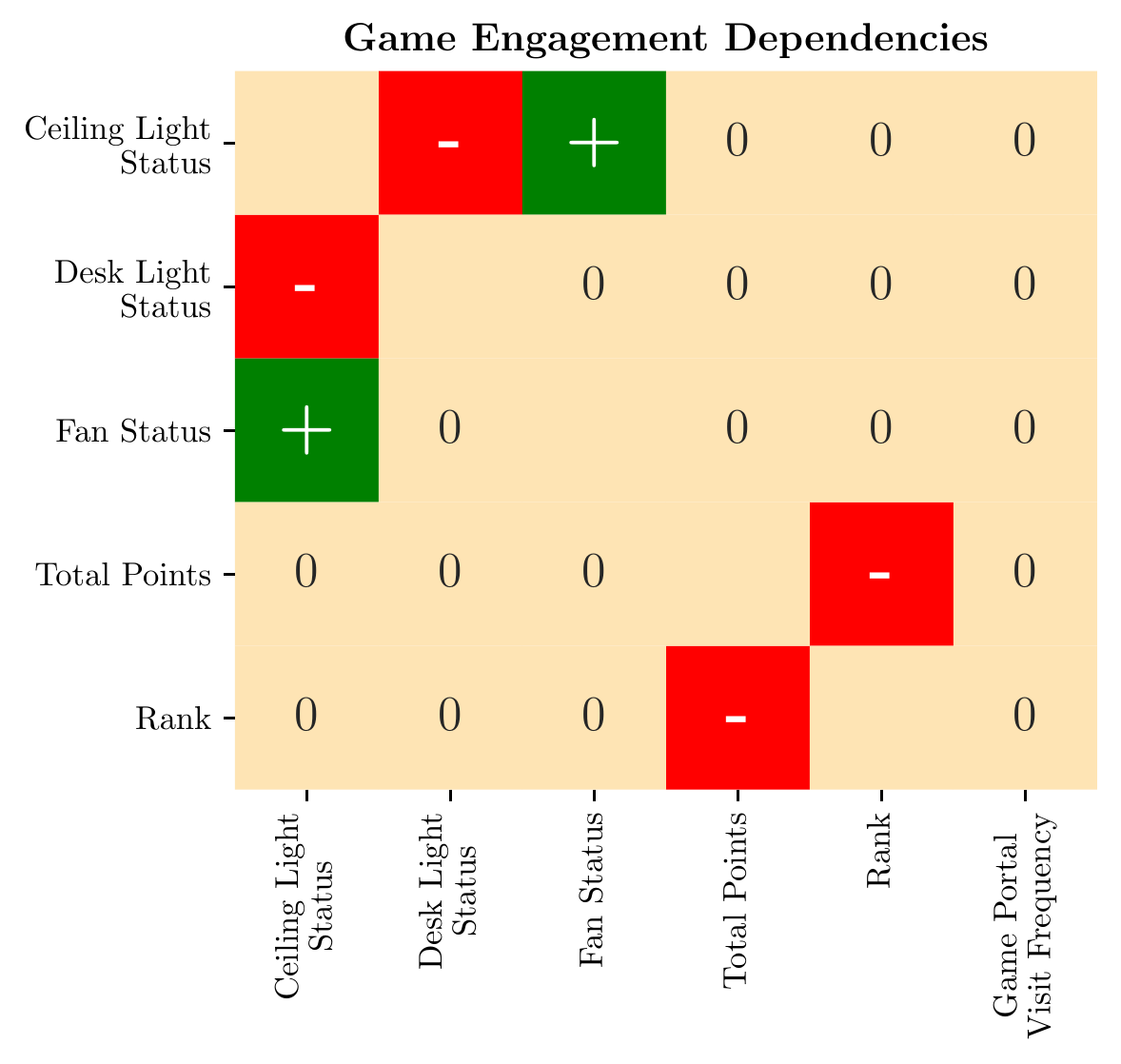}
    }
\caption{{\color{black}Feature correlations for a High Energy Efficient Player ($\in$ $C^{High}$)}}\label{fig_high}
\end{figure}
increase during the breaks and finals, given by positive correlation with total points, but it is not significant as compared to other players during the same period, thus increasing the rank. External parameters play a significant role in energy usage behavior of this class. The operation of the ceiling fan is driven by external humidity as given in Figure~\ref{fig_low}(b). Figure~\ref{fig_low}(c) indicates that their frequency of visits to the game web portal is motivated by sub-optimal performance in the game.

Feature correlations for a medium energy efficient player is given in Fig~\ref{fig_med}. The player showcases predictable behaviors with correlations between desk light, ceiling light and ceiling fan usage (Figure~\ref{fig_med}(a)). The player co-optimizes the usage by alternating the use of ceiling and desk light. Different occasions like break, midterm and final are marked by energy saving patterns. Unlike a low energy efficient player, the player in this class tries to save energy in a conscious manner shown by reduced fan usage during the morning and reduced light usage during the afternoon. The fan usage is influenced by the external humidity, shown by Fig~\ref{fig_med}(b). The game engagement patterns for a player in this class (Fig~\ref{fig_med}(c)) is similar to that of the low energy efficient class.

Fig~\ref{fig_high} shows the feature correlations for a high energy efficient player. This player also exhibits predictable behavior. Opportunistically, this player saves energy during breaks and midterms as shown by negative correlation between the corresponding flags and rank in Figure~\ref{fig_high}(a). Notice that there exists a negative correlation between midterm flag and total points, indicating decrease in absolute amount of points. However, the points are still higher than the points by other players which marks improvement in the rank. This behavior is completely opposite to what is  exhibited by a player in low energy efficient class. The player is neither affected by the time of the day, nor by the external factors (Figure~\ref{fig_high}(b)) showing a dedicated effort to save energy. The game engagement behavior, given in Figure~\ref{fig_high}(c) is inconclusive, possibly due to dominance by other energy saving factors.

\subsubsection{Causal Relationship between features}
To enhance the explainable nature of our model, we studied the causal relationship between features using Granger causality test. Granger causality is a statistical test used to determine causal relationship between two signals. If signal A granger-causes signal B, then past values of A can be used to predict B for future timesteps beyond what is available for B. The results for causal relationship study is given in Table~\ref{tab:grangers_causality}. Under null hypothesis $H_0$, $X$ does not Granger-cause $Y$. So, a p-value lower than $0.05$ (5\% significance level) indicates a strong causal relationship between the tested features and implies rejecting the null hypothesis $H_0$.

The p-values (shaded in blue) for which Granger causality is established are highlighted in the table. Interestingly, for medium and high energy efficient building occupants, ceiling fan usage causes ceiling light usage. This in fact confirms the predictive behavior for them as mentioned earlier. In both low and medium energy efficient building occupants, external humidity causes ceiling fan usage. This is an indicator that their energy usage is affected by external weather conditions. However, for high energy efficient building occupants external humidity doesn't cause ceiling fan usage. This shows that they are highly engaged with the proposed gamification interface and try to minimize their energy usage. Another interesting result is that the evening label causes ceiling light usage for both low and medium energy efficient building occupants. But this is not the case for high energy efficient building occupants, for whom ceiling light usage is better optimized as a result of their strong engagement with the ongoing social game, leading to exhibition of better energy efficiency.
\begin{table*}[t]
  \centering
  \setlength\arrayrulewidth{0.8pt}
  \resizebox{\columnwidth}{!}{
  \begin{tabular}{|c|c|c|c|c|c|c|c|c|c|c|c|c|c|c|c|}
  \rowcolor{Gray}
  \hline
    \multicolumn{1}{|c|}{Test whether $X$ causes $Y$} & \multicolumn{2}{c|}{Fan $\Rightarrow$ Ceiling Light} & \multicolumn{2}{c|}{Humidity $\Rightarrow$ Fan} & \multicolumn{2}{c|}{Desk Light $\Rightarrow$ Fan} & \multicolumn{2}{c|}{Ceiling Light $\Rightarrow$ Desk Light} & \multicolumn{2}{c|}{Morning $\Rightarrow$ Desk Light} & \multicolumn{2}{c|}{Afternoon $\Rightarrow$ Fan} & \multicolumn{2}{c|}{Evening $\Rightarrow$ Ceiling Light} \\ \hline
    \rowcolor{Gray}
Player type & p-value & F-statistic & p-value & F-statistic & p-value & F-statistic & p-value & F-statistic & p-value & F-statistic & p-value & F-statistic & p-value & F-statistic\\ \hline
Low Energy Efficient & 0.54 & 0.37 & \cellcolor{blue!25}\textbf{0.004} & 8.12 & 0.06 & 3.55 & 0.81 & 0.06 & 0.4 & 0.71 & \cellcolor{blue!25}\textbf{0.01} & 6.1 & \cellcolor{blue!25}\textbf{0} & 25.3\\ \hline
Medium Energy Efficient & \cellcolor{blue!25}\textbf{0} & 21.2 & \cellcolor{blue!25}\textbf{0.008} & 7.06 & \cellcolor{blue!25}\textbf{0} & 113.6 & \cellcolor{blue!25}\textbf{0} & 25.8 & 0.23 & 1.41 & 0.46 & 0.55 & \cellcolor{blue!25}\textbf{0.0007} & 11.5\\ \hline
High Energy Efficient & \cellcolor{blue!25}\textbf{0} & 21.9 & 0.12 & 2.36 & 0.99 & 0.003 & 0.93 & 0.007 & 0.63 & 0.22 & \cellcolor{blue!25}\textbf{0.04} & 4.2 & 0.52 & 0.41\\ \hline
\end{tabular}}
\caption{Causality test results among various potential causal relationships. In bold are the p-values (shaded in blue) in cases that Granger causality is established through F-statistic test between features. p-values lower than $0.05$ indicate strong causal relationship in 5\% significance level}
\label{tab:grangers_causality}
\end{table*}

\subsection{Energy Savings Through Gamification}

\begin{figure}[h]
    \subfloat[A/C daily average usage vs. baselines]{%
    \includegraphics[width=0.492\textwidth]{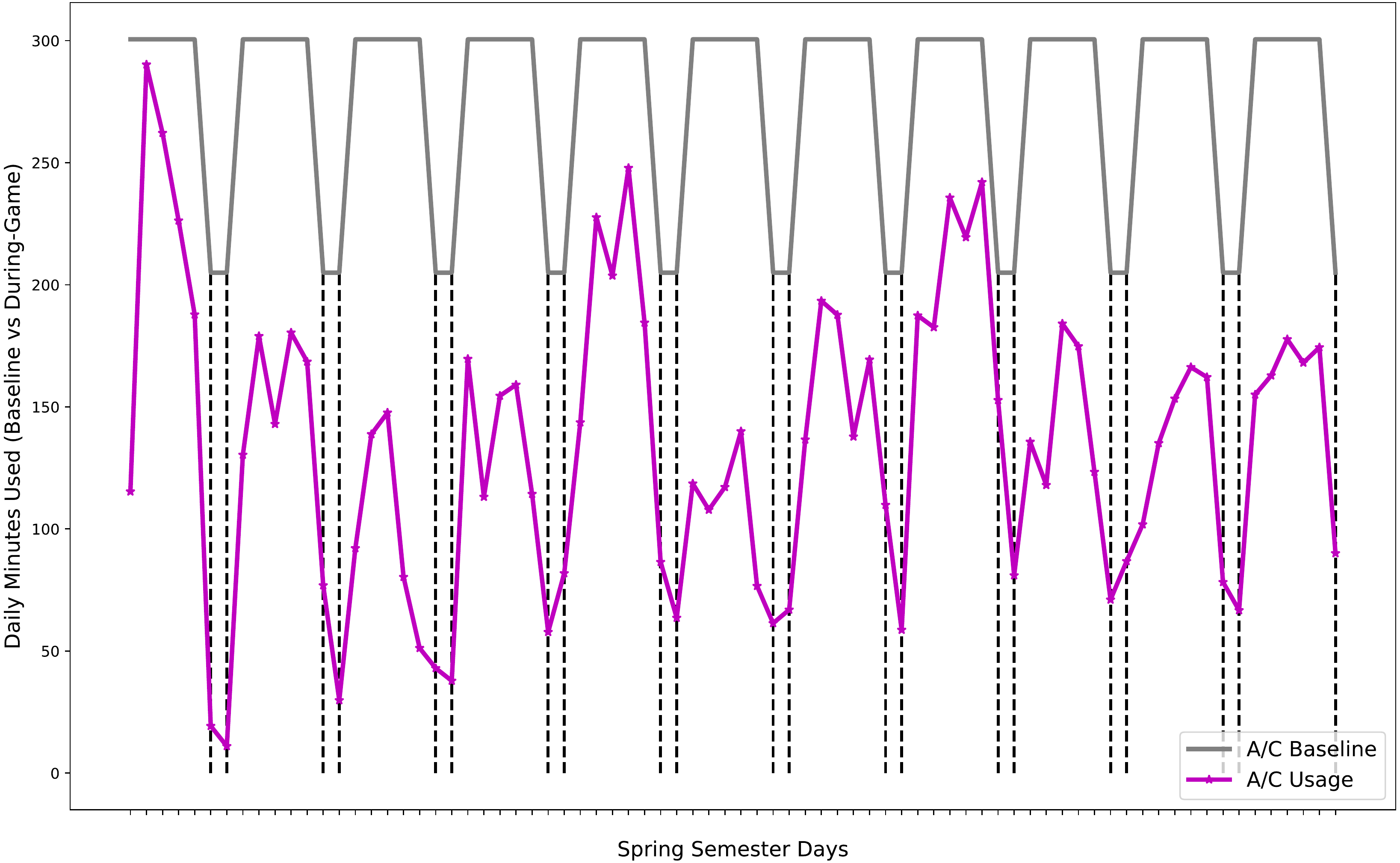}
    }
    \hfill
    \subfloat[Ceiling fan daily average usage vs. baselines]{%
    \includegraphics[width=0.49\textwidth]{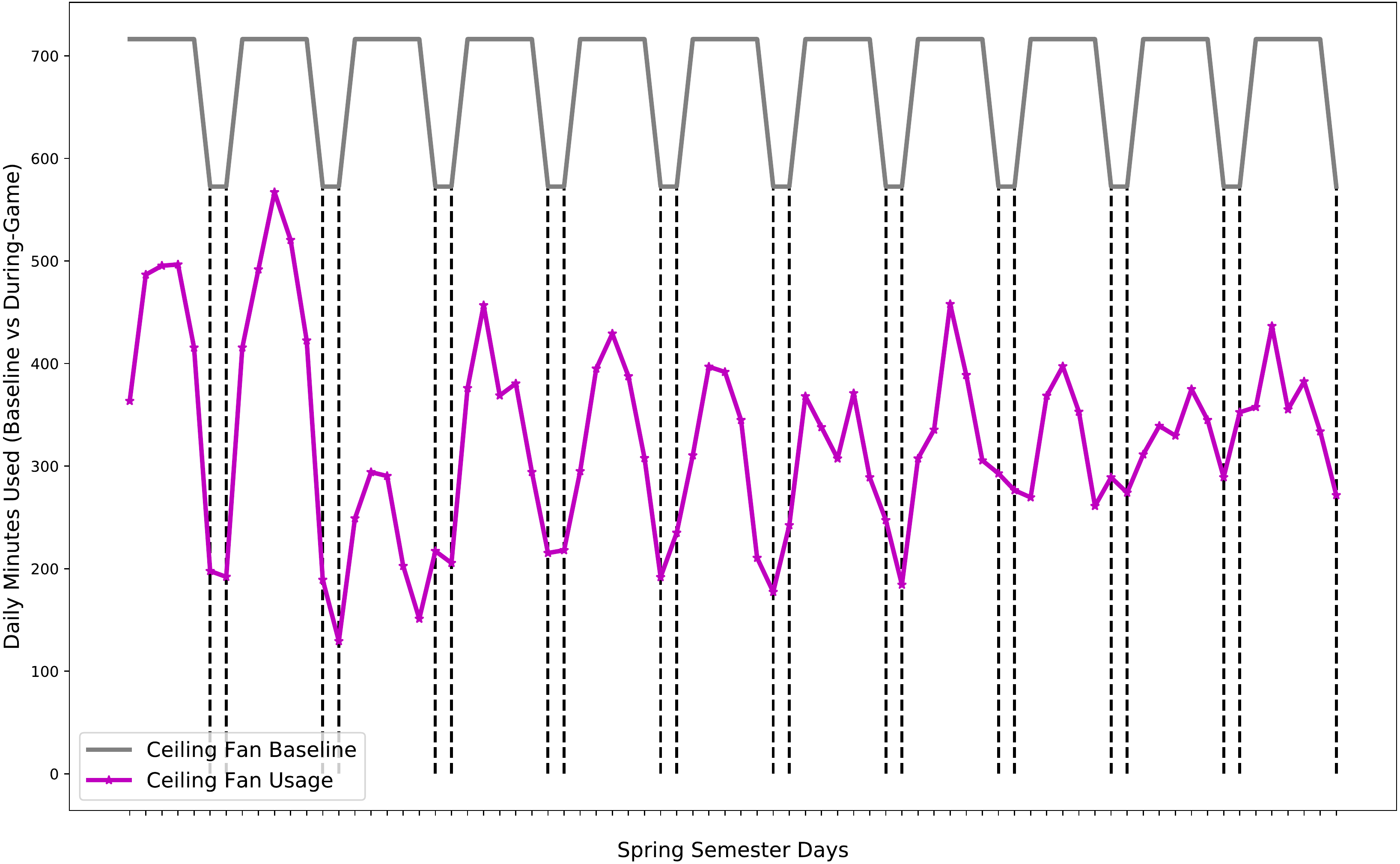}
    }
\caption{Spring semester daily average minutes usage compared to weekday and weekend average baselines. Vertical black dashed lines indicate a weekend period.}
\label{fig:Spring_Energy}
\end{figure}

\begin{table}[ht!]
\centering
\setlength\arrayrulewidth{0.5pt}
\tiny
\caption{Hypothesis testing for Fall \& Spring Game (before vs. after) from minutes per day usage.}
\label{tab:hyp_all}
\begin{tabular}{|c|c|c|c|c|c|c|c|c|}
\hline
\rowcolor{Gray}
   Fall Semester & \multicolumn{4}{c|}{Weekday} & \multicolumn{4}{c|}{Weekend}\\
 \hline
 \rowcolor{Gray} Device & Before (Mean) & After (Mean) & $p$-value & $\Delta$ \% & Before (Mean) & After (Mean)  & $p$-value & $\Delta$ \%\\
 \hline
Ceiling Light & 417.5 & 393.9 & 0.02 & 5.6 & 412.3 & 257.5 & 0 & 37.6\\
 \hline
 Desk Light & 402.2 & 157.5 & 0 & 60.8  & 517.6 & 123.3 & 0 & 76.2\\
 \hline
Ceiling Fan & 663.5 & 537.6 & 0 & 19.0  & 847.1 & 407.0 & 0 & 51.9\\ 
\hline
\hline
\rowcolor{Gray}
   Spring Semester & \multicolumn{4}{c|}{Weekday} & \multicolumn{4}{c|}{Weekend}\\
   \hline
 \rowcolor{Gray} Device & Before (Mean) & After (Mean) & $p$-value & $\Delta$ \% & Before (Mean) & After (Mean)  & $p$-value & $\Delta$ \%\\
 \hline
 Ceiling Light & 452.0 & 314.2 & 0 & 30.5 & 426.0 & 195.6 & 0 & 54.1\\ 
 \hline
 Desk Light & 430.1 & 104.6 & 0 & 75.7 & 509.4 & 81.5 & 0 & 84\\
 \hline
Ceiling Fan & 777.4 &  541.6  & 0 & 30.3 & 847.1 & 331.8 & 0 & 60.8\\ 
\hline
 Air Con & 469.8 & 225.8 & 0 & 51.9 & 412.3 & 81.8 & 0 & 80.2\\
 \hline
\end{tabular}
\end{table}

In this section, we present the resulting energy savings in both Fall and Spring semester versions of the social game. Our gamification framework introduces occupants to a friendly non-cooperative game and motivates reductions in their energy usage. Through the deployed IoT sensors and custom web portal, each individual building occupant received live feedback about their room's usage and their energy efficiency throughout the day. In Figure~\ref{fig:Spring_Energy}, we present the average daily usage in minutes compared to average weekday and weekend baselines. The vertical black dashed lines indicate a weekend period, which has a different corresponding average baseline target for the occupants. In terms of energy usage and savings, A/C and ceiling fan resources demonstrate an impressive reduction in usage, especially during weekends.

For quantifying the results, we employ hypothesis testing (A/B testing) using dorm occupant usage data before and after the beginning of the experiment. In Table~\ref{tab:hyp_all}, we see the hypothesis testing values for the different devices in both iterations of the experiment (Fall and Spring). In this table, the "before" column denotes the data points gathered from before the game was officially started, while "after" is during the game period. Data points in the tables are bucketed in both weekday and weekend data and represent the average usage of all of the occupants. Usage is defined in minutes per day. In all of the devices, we have a significant drop in usage between the two periods. Drop in usage is given in the column named $\Delta$ \%, and indicates reduction in the average usage of all of the participating occupants. The p-values resulting from the 2-sample t-tests show that the change in usage patterns is significant. Furthermore, we can see a much larger drop in usage is achieved over the weekends. These results are significant in that they demonstrate the capacity of our methods to optimally incentivize occupants in residential buildings to make their energy usage more efficient.

\section{Conclusion}
\label{conclusion}

This study presents a general framework for utility learning in sequential decision-making models. We leveraged several Deep Learning architectures and proposed a novel sequential Deep Learning classifier model. We also introduced a framework that serves as a basis for creating generative models, which are ideal for modeling and simulating human-building interaction toward improving energy efficiency. To demonstrate the utility learning methods, we applied them to data collected from a smart building social game where the goal was to have occupants optimize their room's resources. We were able to estimate several agent profiles and significantly reduce the forecasting error compared to all benchmark models. The deep sequential utility learning framework outperformed all other models being considered, and it improved prediction accuracy to an extraordinary degree in specific examples. We also performed explainability analysis using graphical lasso and granger causality to confirm the data encoded knowledge on human decision making towards energy usage in competetive environments.

This last result shows that a Deep Learning architecture that handles a sequential data process has the effect of improving the overall accuracy. In this application we apply these methods specifically to smart building social game data; however, it can generalize to other scenarios with the task of inverse modeling of competitive agents, and it provides a useful tool for many smart infrastructure applications where learning decision-making behavior is crucial. Under our gamification application, occupants were highly motivated to drastically reduce their energy impact. This result is even more significant considering the fact that no effort was directed towards optimizing the incentive design for encouraging energy efficient behavior. Hence, research in optimal incentive design mechanisms should be pursued in the context of this work. Recent research on segmentation analysis using graphical lasso~\citep{das2019novel} can further improve the intelligent incentive design process. Furthermore, special attention should be given to the management of pricing and how it affects the dynamics between the smart building and utility provider for applications like demand response programs. In general, we have demonstrated that our proposed framework can be used successfully for the purpose of accurately forecasting energy usage. However, deep learning models require a continuous feed of data and are not particularly robust to missing data points. This poses a challenge to many real-world applications, especially in such cases where there might be missing data due to IoT sensors losing connection. Hence, we identify this as a limitation that should be addressed for effective implementation of deep learning models in energy game-theoretic framework. Despite these constraints, we have shown that our implementation of a gamification approach to human-building interaction in smart infrastructure offers tremendous opportunities for improving energy efficiency and smart grid management.

\section{Acknowledgement}

The authors would like to thank Chris Hsu, the applications programmer at CREST laboratory, who developed and deployed the web portal application as well as the social game data pipeline architecture. Also, we want to thank Energy Research Institute (ERI@N) at Nanyang Technological University. Geraldine Thoung, Patricia Alvina and Nilesh Y. Jadhav at ERI@N kindly supported and helped during the social game experiment. This work was supported by the Republic of Singapore’s National Research Foundation through a grant to the Berkeley Education Alliance for Research in Singapore for the Singapore–Berkeley Building Efficiency and Sustainability in the Tropics (SinBerBEST) Program. The work of I. C. Konstantakopoulos was supported by a scholarship of the Alexander S. Onassis Public Benefit Foundation.

\bibliography{social_game}
\bibliographystyle{iclr2020_conference}
\clearpage
\section*{Appendix}
\noindent \textbf{A. Description of Graphical Lasso algorithm}\\
\begin{tikzpicture}[yscale=3] 
\draw [line width=0.65mm, black ] (0,-1) -- (14,-1) node [right]{};;
\end{tikzpicture}
 Algorithm 1: GRAPHICAL LASSO ALGORITHM FOR GAUSSIAN GRAPHICAL MODELS\\
\begin{tikzpicture}[yscale=3] 
\draw [line width=0.65mm, black ] (0,0) -- (14,0) node [right]{};;
\end{tikzpicture}
\begin{enumerate}[1.]
\item For vertices $h = 1, 2, \cdots ,H$:
\begin{enumerate}[a.]
    \item Calculate initial loss ${\|Y_{h} - Y^T_{V\backslash h}\beta^{h}\|}^{2}_{2}$
    \item Untill Convergence:
    \begin{enumerate}[i.]
        \item Calculate partial residual $r^{(h)}$ = $Y_h$ - $Y^T_{V\backslash h}\beta^{h}$
        \item For all $j \in {V\backslash h}$, Get ${\beta}^{h,new}_j$ = ${\textbf S_{\lambda}\big(\frac{1}{N}\langle r^{(h)},Y_j\rangle\big)}$ 
        \item Compute new loss = ${\|Y_{h} - Y^T_{V\backslash h}\beta^{h,new}\|}^{2}_{2}$
        \item Update ${\beta^{h}}$ = ${\beta}^{h,new}$
    \end{enumerate}
    \item Get the neighbourhood set $\mathcal{N}(h)= supp({{\beta}}^{h})$ for $h$
\end{enumerate}
\item Combine the neighbourhood estimates to form a graph estimate ${G} = (V, \textit{E}$) of the random variables.
\end{enumerate}
\begin{tikzpicture}[yscale=3] 
\draw [line width=0.65mm, black ] (0,-1) -- (14,-1) node [right]{};;
\end{tikzpicture}

$\textbf S_{\lambda}(\theta)$ is soft thresholding operator as $sign(\theta)(|\theta|-\lambda)_{+}$.\\
\hspace{50mm}

For optimal design of penalty factor $\lambda$ in Graphical Lasso run for a vertex s, we take 10 values in logarithmic scale between $\lambda_{max}$ and $\lambda_{min}$ as given below and conduct a line search to find the penalty factor which brings the minimum loss.
\begin{align}
 \lambda_{max} = \frac{1}{N}\underset{j\in V\backslash h}{max}|\langle Y_j,Y_h\rangle|, \quad \lambda_{min} = \frac{\lambda_{max}}{100}
 \end{align}
Implementing a coordinate descent approach ~\cite{mainbook}, the time complexity of the proposed algorithm is $O(HN)$ for a complete run through all $H$ features. We also do 5-fold cross validation to ensure accurate value of the coefficients $\beta^{h}$. Use of partial residuals for each node significantly reduces the time complexity of the algorithm.

\end{document}